\documentclass[preprint,twoside,12pt]{elsarticle}

\usepackage[dvipsnames]{xcolor}
\usepackage{array}
\usepackage{tabularx}
\usepackage{hyperref}
\usepackage{amsfonts} 
\usepackage{breqn}
\usepackage{multirow}
\usepackage{pdflscape}
\usepackage{float}
\usepackage{caption}
\usepackage{mathrsfs}
\usepackage{bm}   
\usepackage{amsmath}
\usepackage{caption}
\usepackage{amsmath, amsthm, amssymb}
\usepackage{fancyhdr}
\usepackage{appendix}
\usepackage[a4paper, left=2.5cm, right=2.5cm, top=3.5cm, bottom=2.5cm]{geometry}
\newcommand{\EvenLeft}{}
\newcommand{\EvenRight}{}
\newcommand{\OddLeft}{}
\newcommand{\OddRight}{}

\biboptions{sort&compress}
\makeatletter
\newcommand{\PreAbstractNote}{%

}

\patchcmd{\pprintMaketitle}{\hrule\vskip12pt}{\PreAbstractNote\hrule\vskip12pt}{}{}
\patchcmd{\MaketitleBox}{\hrule\vskip12pt}{\PreAbstractNote\hrule\vskip12pt}{}{}
\makeatother

\usepackage[ruled,vlined]{algorithm2e}
\hypersetup{
 colorlinks,
 citecolor=black,
 filecolor=black,
 linkcolor=black,
 urlcolor=black
}
\usepackage{amssymb}

\journal{Neural Networks}

\DeclareCaptionLabelFormat{myformat}{\textbf{Fig. #2}}
\theoremstyle{definition}
\newtheorem{definition}{Definition}[section]
\begin{document}

\begin{frontmatter}

\title{Learning in PINNs: Phase Transition, Diffusion Equilibrium, \\ and Generalization\tnoteref{pubnote}}
\author[inst1]{Sokratis J. Anagnostopoulos}
\author[inst2]{Juan Diego Toscano}
\author[inst1]{\\ Nikolaos Stergiopulos}
\author[inst2]{George Em Karniadakis}

\affiliation[inst1]{organization={Laboratory of Hemodynamics and Cardiovascular Technology, EPFL},
 city={Lausanne},
 postcode={1015}, 
 state={VD},
 country={Switzerland}}

\affiliation[inst2]{organization={Division of Applied Mathematics, Brown University},
 city={Providence},
 postcode={02912}, 
 state={RI},
 country={USA}}

\affiliation[inst3]{organization={School of Engineering, Brown University},
 city={Providence},
 postcode={02912}, 
 state={RI},
 country={USA}}

\tnotetext[pubnote]{Published in \textit{Neural Networks}: \href{https://doi.org/10.1016/j.neunet.2025.107983}{doi:10.1016/j.neunet.2025.107983}. \\ 
\hspace*{1.5em} Updated to match publication; added appendix on batch-independent SNR.}

\begin{abstract}

{We investigate the learning dynamics of fully-connected neural networks through the lens of the neural gradient signal-to-noise ratio (SNR), examining the behavior of first-order optimizers in non-convex objectives. Interpreting the drift/diffusion phases as proposed in the information bottleneck theory, we identify a third phase termed “diffusion equilibrium” (DE), a stable training phase characterized by highly-ordered neural gradients across the sample space. This phase is marked by an abrupt transition, where sample-wise gradients align (SNR increases), and stable optimizer convergence. Moreover, we find that when homogeneous residuals are also met across the sample space during the DE phase, this leads to better generalization, as the optimization steps are equally sensitive to each sample. Based on this observation, we propose a sample-wise re-weighting scheme, which considerably improves the residual homogeneity and generalization in quadratic loss functions, by targeting the problematic samples with large residuals and vanishing gradients. Finally, we explore the information compression phenomenon, pinpointing a significant saturation-induced compression of activations at the DE phase transition, driven by the sample-wise gradient directional alignment. Interestingly, it is during the saturation of activations that the model converges, with deeper layers experiencing negligible information loss. Supported by experimental examples on physics-informed neural networks (PINNs), which highlight the critical role of gradient agreement due to their inherent PDE-based interdependence of samples, our findings suggest that when both sample-wise gradients and residuals are ordered, this leads to faster convergence and better generalization. Identifying phase transitions could improve deep learning optimization strategies, enhancing physics-informed methods and machine learning performance.}

\end{abstract}


\end{frontmatter}

\newpage

\section{Introduction}
\subsection{Phase transitions in deep learning}
The optimization process in deep learning can vary significantly in terms of smoothness and convergence rate, depending on various factors such as the complexity of the model, the quality/quantity of the data or the loss landscape characteristics. However, for non-convex objectives this process has often been observed to be far from smooth and steady,  often dominated by discrete, successive phases. Recent studies have shed light on several key aspects influencing these phases and the overall optimization dynamics \cite{zhou2019toward,smith2019super,lewkowycz2020large,lee2019wide,katsnelson2021self,cohen2020gradient,cohen2022adaptive,damian2022self,kosson2023rotational,nakkiran2021deep}.

The importance of gradient noise in escaping local optima of non-convex optimization has been explored, demonstrating its role in guaranteeing polynomial time convergence to a global optimum \cite{zhou2019toward}. The authors of the same work suggest the existence of a phase transition for a perturbed gradient descent (GD) algorithm, from escaping local optima to converging to a global solution as the artificial noise decreases. In a later work, a phenomenon called ``super-convergence" has been highlighted, where models trained with a two-phase cyclical learning rate may lead to improved regularization balance and generalization \cite{smith2019super}. Furthermore, recent investigations have discovered a two-phase learning regime for full-batch GD, characterized by distinct behaviors \cite{lewkowycz2020large}. During the ``lazy” phase ($\eta<2/\lambda_0$), the model behaves linearly about its initial parameters \cite{lee2019wide}. However, optimal performance is usually achieved during the “catapult” phase ($2/\lambda_0<\eta<\eta_{max}$), which exhibits instabilities due to the increased curvature. The findings of this study are also supported using the NTK method \cite{jacot2018neural}, which has been proven to be an effective tool for studying deep learning dynamics in the infinite width limit. Other researchers have demonstrated that neural networks tend towards a self-organized critical state, characterized by scale invariance, where both trainable and non-trainable parameters are driven to a stochastic equilibrium \cite{katsnelson2021self}, which is also present in many biological systems.

An intriguing study showed empirically that contrary to the theoretical clues on stability, GD very often trains at the ``edge of stability”, which defines a regime of hypercritical sharpness (max Hessian eigenvalue), that hovers just above $2/\eta$, during which there is non-monotonic convergence \cite{cohen2020gradient}, present also for adaptive optimizers \cite{cohen2022adaptive}. The authors also posed questions regarding the mystery of progressive sharpening preceding the ``edge of stability” phase. A follow-up paper explains this equilibrium state using the cubic Taylor expansion, showing that GD has an inherent self-stabilizing mechanism, which decreases the sharpness whenever extreme oscillations are present \cite{damian2022self}. Analyzing the dynamics of popular optimizers, such as Adam \cite{kingma2014adam}, has also revealed insights into the non-convex learning process. For instance, the concept of rotational equilibrium studies a steady state of angular network parameter updates, offering a deeper understanding of the phase transitions experienced by the weight vector during optimization \cite{kosson2023rotational}. Additionally, the ``double-descent" phenomenon has emerged as a notable discovery, revealing a critical range of network width that negatively impacts generalization performance, with implications for the success of large language models \cite{nakkiran2021deep}, while this transition can also be met epoch-wise. Collectively, these findings emphasize the complexity of deep learning optimization and the existence of discrete learning phases, offering valuable insights for improving optimization algorithms and understanding deep neural network behavior. A reasonable speculation could be that most of these phase transitions largely coincide, but this investigation is beyond the scope of our study.

\subsection{Physics-informed learning}

Physics-informed neural networks (PINNs) have emerged as an alternative to classical numerical methods for solving general non-linear partial differential equations (PDEs) for forward and inverse problems \cite{raissi2019physics}. Unlike traditional methods, which use numerical integration to propagate the solution from the initial or boundary conditions, PINNs approximate the PDE solution through a gradient-based optimization process. This optimization involves minimizing a constrained loss function designed to concurrently satisfy the governing physical laws and initial/boundary conditions, while fitting any available experimental data (inverse solution) \cite{raissi2020hidden,boster2023artificial,oommen2022solving,cai2021artificial,cao2024deep}.

PINNs have gained popularity due to their superiority in inverse problems, motivating researchers to develop multiple methods to ease the optimization process. Some of these studies focus on the network architecture via domain decompositions \cite{kopanivcakova2023enhancing}, architecture reformulations \cite{wang2021understanding, salimans2016weight,wang2024piratenets,jiang2024densely, zhang2023artificial}, input dimension expansions \cite{guan2022dimension,wang2021eigenvector}, sequential learning \cite{krishnapriyan2021characterizing,wight2020solving}, or adaptive activation functions \cite{jagtap2020adaptive}. Similarly, other studies attempt to simplify the problem by reducing the number of loss terms via exact boundary imposition \cite{sukumar2022exact,leake2020deep,lu2021physics} or PDE reformulations \cite{jin2021nsfnets,basir2022investigating}. 

On the other hand, other approaches have focused on the imbalances of the constrained optimization. Maintaining a global balance (i.e., between each loss term) and local balance (i.e., between training samples) is crucial to ensure that all constraints are adequately minimized across the domain. Some researchers have attempted to deal with these imbalances by adaptively resampling points in critical areas \cite{lu2021deepxde,wu2023comprehensive,zapf2022investigating,deguchi2023dynamic} while other studies involve re-weighting methods, which assign adaptive global or local multipliers. In general, global multipliers modify the average contribution of each loss term ~\cite{wang2021understanding,wang2024respecting,xiang2022self,liu2021dual} while local multipliers adjust the local contribution of each training sample~\cite{anagnostopoulos2024residual,mcclenny2020self,basir2022investigating,basir2022physics,basir2023adaptive,zhang2023dasa,son2023enhanced,wang2024respecting}. 

\subsection{Information bottleneck method}

The information bottleneck (IB) method provides an interpretation of neural network training and performance from the information theory perspective. It formulates a methodology for estimating the optimal trade-off between compression and prediction in supervised learning, and constructs an idealized principle for creating a compressed or ``bottlenecked" representation of the layer activations $\mathcal{T}$ for an input variable $\mathcal{X}$, that retains as much information as possible about an output variable $\mathcal{Y}$ \cite{tishby2000information,tishby2015deep}. The theory utilizes the concept of mutual information $I(\mathcal{X},\mathcal{Y})$, which measures the amount of information obtained about $\mathcal{Y}$ by observing $\mathcal{X}$, suggesting that the optimal model representations retain all the relevant information about the output while discarding irrelevant details about the input, hence creating an ``information bottleneck". A remarkable consequence of IB is that deep learning is governed by two discrete phases: the fitting and the diffusion \cite{shwartz2017opening,goldfeld2020information,shwartz2022information}, which are separated by a phase transition, captured by the signal-to-noise ratio (\text{SNR}) of the gradients. The theory suggests that most of the learning occurs during the slow diffusive phase, where the model is able to generalize well.

Although the IB theory has remained a subject of increasing interest over the years, there have been ongoing debates regarding some of its claims, mainly the compression phenomenon and its contribution to generalization \cite{saxe2019information,alomrani2021critical,goldfeld2018estimating,guo2023ib,lorenzen2021information,basirat2021geometric,geiger2021information}. In addition, providing an accurate estimate of the mutual information is non-trivial, with studies that focus on different methodologies and how they affect the interpretation of the process \cite{saxe2019information,lorenzen2021information}. {It has been pointed out that since the information is infinite due to the continuous nature of DNNs, quantifying it heavily depends on the quantization method \cite{lorenzen2021information,geiger2021information}. Moreover, this study showed that the compression phase (decreasing $I(\mathcal{X},\mathcal{Y})$) is not as common as initial claims suggest. A recent study provides a clearer view of quantifying $I(\mathcal{X},\mathcal{Y})$, and supports that the compression mechanism is driven by progressive clustering of each layer’s activations, which is also present in deterministic DNNs \cite{goldfeld2018estimating}. The same authors justify the binning approximation as a measure of clustering and suggest that compression is not necessarily related to generalization. These results validate that when compression is present, it does not arise from gradient stochasticity but rather due to geometric clustering for models trained with stochastic or full batch GD \cite{lorenzen2021information,goldfeld2018estimating}. In general, many recent studies find that there is no evident causal connection between compression and generalization, implying that networks which do not compress are still capable of generalizing \cite{saxe2019information,goldfeld2018estimating,geiger2021information}.}The IB theory has also been explored for scientific machine learning (ML), for an IB-inspired architecture of uncertainty quantification for PINNs and DeepONets \cite{guo2023ib}.
\begin{figure}[H]
 \centering
 \includegraphics[width=.85\textwidth]{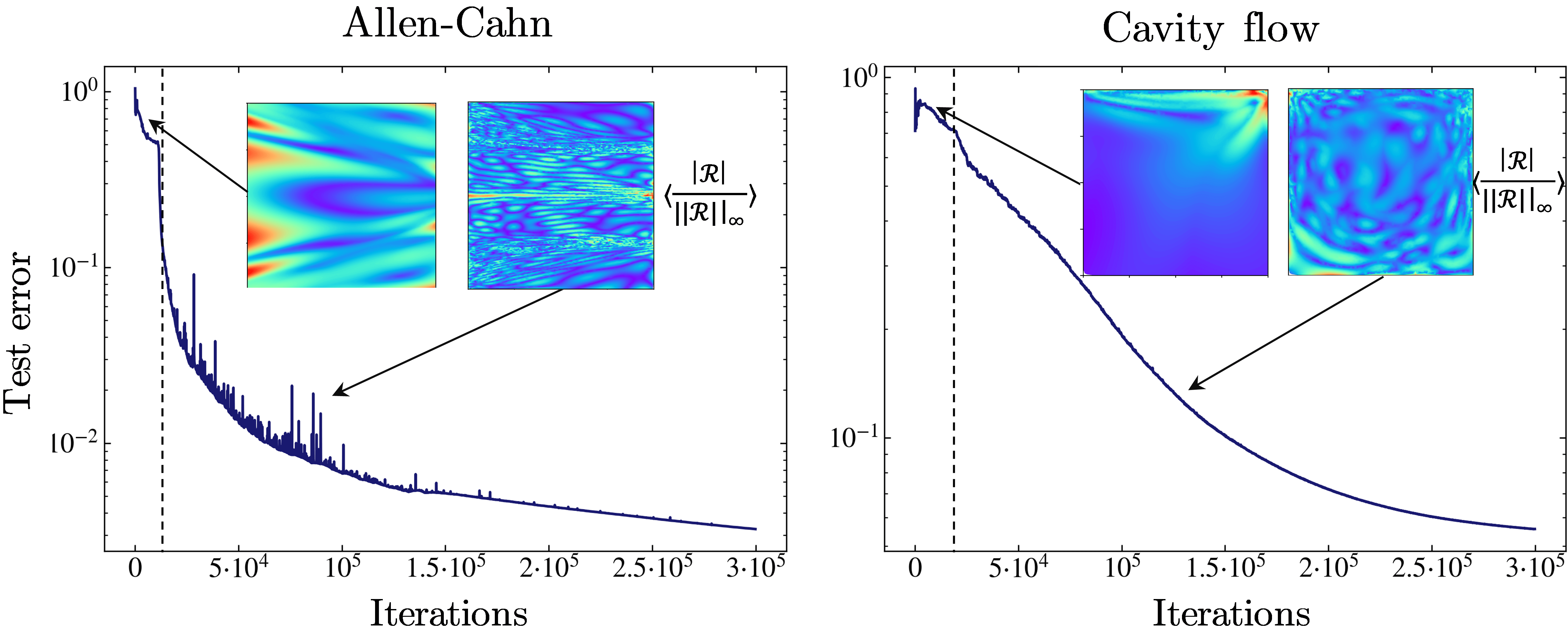}
 \caption[Phase transition in PINNs]{ \textbf{Phase transition in PINNs:} The test error between the prediction and the exact solution converges faster after DE (dashed lines), which occurs with an abrupt phase transition. Although the convergence starts during the onset of the diffusion phase, the optimal training performance is achieved at the equilibrium when the gradients of different batches become equivalent with homogeneous residuals, indicating a general agreement on the direction of the optimizer steps.}
 \label{p3:fig0}
\end{figure}

\subsection{Main contributions}

The aim of this work is to contribute to a better understanding of phase transitions in deep learning, with a focus on PINNs applications. In the Methods, we analyze the training dynamics of full-batch GD using the gradient \text{SNR} and demonstrate how the equilibrium of the iteration-wise \text{SNR} used in Adam does not guarantee uniform learning across the sample space, which can lead to instabilities, overfitting and bad generalization. This weakness caused by the quadratic loss averaging can have detrimental effects on PINNs, where the samples (collocation points) should ideally converge simultaneously to achieve a low test error.

We provide novel evidence for the existence of phase transitions proposed by the IB theory (fitting/diffusion) for PINNs trained with Adam, and reveal a third phase at the end of the diffusion process which we call ``diffusion equilibrium” (DE), where stable convergence occurs (Fig. \ref{p3:fig0}). We show that this phase {marks a first-order sample-wise gradient alignment, which, however, does not guarantee residual homogeneity among samples. We argue that when both sample-wise gradients and residuals are highly ordered, this leads to better generalization where there is an equivalence of samples and a stable equilibrium of the optimizer can be reached. This comes as a complementary finding to the common belief that the models learn during a highly stochastic process (also supported by IB theory).} Thus, we propose a re-weighting technique based on the moving average of the residuals named residual-based attention (RBA) \cite{anagnostopoulos2024residual} to further support the hypothesis that homogeneous residuals lead to better generalization. We compare the performance of the vanilla and RBA models, showing that the latter can achieve DE faster, leading to better generalization for most cases. 

Finally, we demonstrate how the DE phase coincides with information compression or saturation, which we interpret as the ``binarization” of the activations. During the abrupt transition into ``DE”, {the weights rapidly increase due to the gradient aligment}, causing saturation of the neuron activations. {Hence, while compression is an expected consequence and not necessarily a prerequisite for generalization.} We also deploy a relative-binning approach and observe that although there is no major information loss for deeper layers, there is a hierarchy for the information carried by each layer, which is consistent with the IB theory. We further demonstrate that it is the middle layers that saturate the most during the saturation process, a clue reminiscent of an encoder-decoder process within the fully connected architecture. {Interestingly, the optimizer converges during the saturation process.}

\section{Methods}
\subsection{Gradient signal and noise for SGD}

Let $\mathcal{X}$ be a dataset of $n$ training samples $x_i|_{i=1}^n$ and $\mathcal{R} (f(x_i, \theta), y_i)$ be the residuals between the prediction of a model $f$ with parameters $\theta$ and the correct labels $y_i$. The minimization problem we aim to solve using the quadratic loss is defined as:

\begin{equation}
 \min_{\theta \in \Theta} \mathcal{L}(\theta) = \frac{1}{n} \sum_{i\in \mathcal{X}} \mathcal{R} (f(x_i; \theta), y_i)^2.
\label{p3:a}
\end{equation}

A common gradient-based optimization method for problems of this form is Stochastic Gradient Descent (SGD), which updates the model parameters at every iteration $t$, using mini-batches $\mathcal{B}$ of size $m$ as:

\begin{equation}
\theta^{t+1} = \theta^t - \eta \cdot \nabla_{\theta}{\mathcal{L}_{\mathcal{B}}},
\label{p3:b}
\end{equation}

\noindent with 

\begin{equation}
\nabla_{\theta}{\mathcal{L}_{\mathcal{B}}} = \frac{1}{m}\sum_{i \in \mathcal{B}} \nabla_{\theta}{\mathcal{R} (f(x_i, \theta), y_i)^2},
\label{p3:c}
\end{equation}

\noindent where $\mathcal{B} \subset \mathcal{X}$ and $\eta$ is the learning rate. Since for uniform sampling each batch gradient $\nabla_{\theta}{\mathcal{L}_{\mathcal{B}}}$ is an unbiased estimator of $\nabla_{\theta}{\mathcal{L}}$, it can be expressed as a composition of the expected gradient and some noise $\epsilon$:

\begin{equation}
\nabla_{\theta}{\mathcal{L}_{\mathcal{B}}} = \nabla_{\theta}{\mathcal{L}} + \epsilon_{\mathcal{B}} = \mathbb{E}[\nabla_{\theta}{\mathcal{L}_{\mathcal{B}}}] + \epsilon_{\mathcal{B}},
\label{p3:d}
\end{equation}

\noindent where $\nabla_{\theta}{\mathcal{L}}$ is the true gradient and $\epsilon_{\mathcal{B}}$ is a zero-mean ($\mathbb{E}[\epsilon_{\mathcal{B}}] = 0$) variable, measuring the deviation of the batch gradient from the true gradient. Thus, Eq. \ref{p3:b} can be expanded as:
\begin{equation}
\theta^{t+1} = \theta^t - \eta \cdot (\mathbb{E}[\nabla_{\theta}{\mathcal{L}_{\mathcal{B}}}] + \epsilon_{\mathcal{B}}),
\label{p3:e}
\end{equation}

\noindent where $\eta \cdot \mathbb{E}[\nabla_{\theta}{\mathcal{L}_{\mathcal{B}}}]$ is the deterministic direction and $\eta \cdot \epsilon_{\mathcal{B}}$ is the stochastic direction that the optimizer will take when it performs a step on each batch $\mathcal{B}$. 

In this study, we measure the generalization performance of the model with the relative $L^2$ error (test error), which is calculated as:

\begin{equation} \textit{Relative } L^2 = \frac{\lVert f(x;\theta) - u_e \rVert_2}{\lVert u_e \rVert_2},
\label{p3:l2}
\end{equation}

\noindent where $f(x;\theta)$ is the prediction of the model and $u_e$ is the exact solution, unseen by the model. For PINNs, the term to be minimized in Eq. \ref{p3:a} is $\mathcal{R} (f(x_i; \theta))^2$, where $\mathcal{R}$ are the PDE residuals. {Note that the relative $L^2$ error is highly sensitive to outliers due to the quadratic norms, therefore to achieve a low test error, the residuals of all samples have to be uniformly low across the sample space.}

\begin{figure}[H]
 \centering
 \includegraphics[width=.9\textwidth]{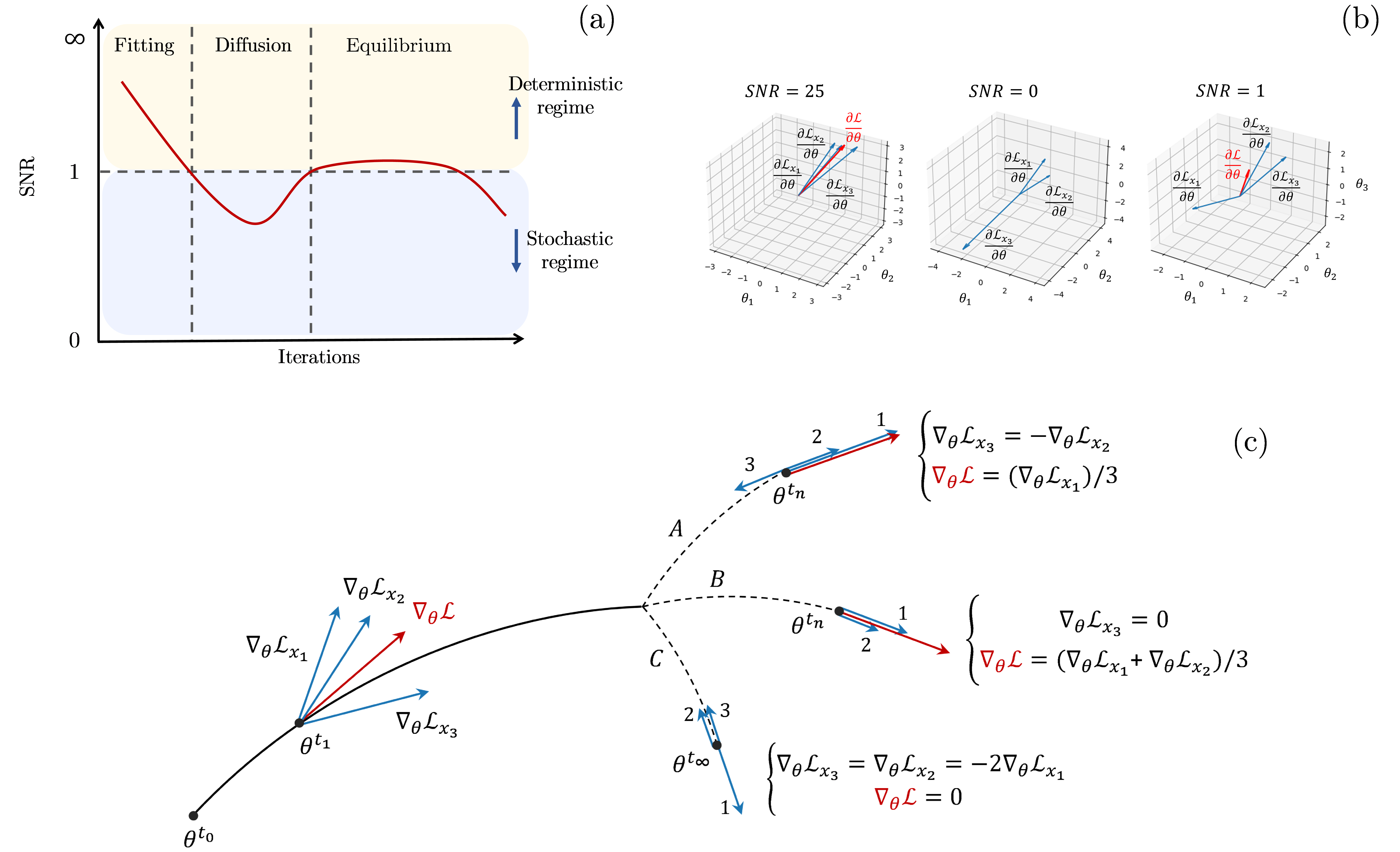}
 \caption[Gradient-based optimization regimes]{ \textbf{(a) Gradient-based optimization regimes:} Indicative $\text{SNR}$ training curve at each full-batch iteration. During the initial learning stages, $\text{SNR} \gg \mathcal{O}(1)$ as the deterministic term dominates with a fast decrease, as the steps become more stochastic. The first two stages of learning are defined as ``fitting" and ``diffusion", {with $\text{SNR}_{fit}>\text{SNR}_{diff}$ and $\frac{\partial}{\partial t}\text{SNR}_{fit} < 0$.} The ``DE" starts when the batch gradients are approximately equivalent, met with an abrupt increase of the \text{SNR}, {where $\text{SNR}_{DE}>\text{SNR}_{diff}$ and $\frac{\partial}{\partial t}\text{SNR}_{DE}\approx 0$.} During the final stages, the \text{SNR} decreases as the signal (numerator) tends to zero and some noise (denominator) persists. \textbf{(b) Batch-wise \text{SNR} directions:} Indicative directions in the parameter space for each \text{SNR} case. For $\text{SNR}>1$, there is an agreement of directions among samples $x_i$, resulting in a deterministic $\frac{\partial\mathcal{L}}{\partial \theta}$ of large magnitude. When $\text{SNR} = 0$, the directions cancel out, indicating convergence to a local minimum. Finally, for $\text{SNR} = 1$, there is balance between determinism and stochasticity in the direction of $\frac{\partial\mathcal{L}}{\partial \theta}$, for which the magnitude is non-zero. \textbf{(c) Training trajectories:} At each step t, Adam aims to maintain a consistent direction based on the gradient signal $\mathbb{E}[\nabla_{\theta}\mathcal{L}] $. However, this does not guarantee agreement in the sample-wise gradients, which can lead to converging in a local minimum where the sample-wise gradients cancel out while being disproportional. Such a scenario could indicate that the model overfits to certain $x_i$ while underfitting to others.}
 \label{p3:regimes}
\end{figure}

To collectively measure the dynamics of all network gradients in a meaningful and efficient way, at each step we split our dataset into batches and calculate the ratio of batch-wise mean ($\mu_j$) and standard deviation ($\sigma_j$) for each parameter $\theta_j$, a metric which is formally defined below.
\begin{definition} Based on the IB theory, the batch-wise signal-to-noise ratio (SNR) quantifies the training dynamics and is defined as:
\end{definition}
\vspace{-1em}
\begin{equation} \text{SNR} = \frac{\lVert \mu \rVert_{2}}{\lVert \sigma \rVert_{2}} = \frac { \lVert \mathbb{E}_{\mathcal{B}}[\nabla_{\theta}{\mathcal{L}_{\mathcal{B}}}] \rVert_{2}} {\lVert std_{\mathcal{B}}[\nabla_{\theta}{\mathcal{L}_{\mathcal{B}}}] \rVert_{2}}
\label{p3:g}
\end{equation}

\noindent where $\lVert \mu \rVert_{2}$ and $\lVert \sigma \rVert_{2}$ are the $L^2$ norms of the batch-wise mean and std of $\nabla_{\theta}{\mathcal{L}_{\mathcal{B}}}$ for all network parameters. The numerator represents the gradient signal, which drives the optimizer to a consistent direction, while the denominator is the gradient noise or variation from the learning signal. Based on IB theory, the task of efficient neural network training lies in the optimal extraction of relevant information (signal) from irrelevant or misleading information (noise). In this work, we study the training dynamics of full-batch Adam trained with the entire dataset $\mathcal{X}$, and calculate $\nabla_{\theta}{\mathcal{L}_{\mathcal{B}}}$ without performing an update of $\theta$ for each $\mathcal{B}$ so that we can investigate the batch-wise behavior on the same iteration $t$. The term which dominates in Eq. \ref{p3:e} defines the optimization regime, which is stochastic for $\text{SNR}<\mathcal{O}(1)$ or deterministic for $\text{SNR}>\mathcal{O}(1)$ (Fig. \ref{p3:regimes}(a)). The IB theory states that neural gradients undergo a phase transition from high to low $\text{SNR}$, which corresponds to a ``fitting phase" of high gradient agreement and a ``diffusion" phase when the gradient fluctuations dominate the signal and achieve a compressed representation. However, in all our case studies, a third phase also emerges for prolonged training (more than $10^4$ iterations), which we call ``DE" (diffusion equilibrium). In this phase, we observe a sudden jump of the $\text{SNR}$ to a stable region due to batch gradient {agreement, where $\text{SNR}_{DE}>\text{SNR}_{diff}$ and $\frac{\partial}{\partial t}\text{SNR}_{DE}\approx 0$. This behavior} implies a high degree of uniformity in the samples and their corresponding sensitivity to the optimizer steps. While the neural network starts converging from the diffusion phase, as shown in previous studies, we suggest that it is after the DE when the neural network can converge optimally. Therefore, diffusion may be considered an exploratory stochastic phase, while the DE phase exploits a consistent direction. Note that a sudden increase in the $\text{SNR}$ is also met in the literature \cite{shwartz2017opening, saxe2019information, shwartz2022information} although it is not explicitly addressed, probably due to the shorter training.

Figure \ref{p3:regimes}(b) shows three indicative cases where $\text{SNR} > 1$, $\text{SNR} = 0$ and $\text{SNR} = 1$, for a network of three parameters $\theta_j$ and three samples $x_i$. When $\text{SNR}>1$, the directions of samples $x_i$ agree, resulting in a deterministic $\nabla_{\theta}{\mathcal{L}}$ of large magnitude. When $\text{SNR} = 0$, the vectors cancel out, indicating convergence. Finally, for $\text{SNR} = 1$, there is an equilibrium between determinism and stochasticity in the direction of $\nabla_{\theta}{\mathcal{L}}$, for which the magnitude is non-zero. Thus, $\text{SNR}$ measures directional and magnitude agreement.

To provide a more intuitive understanding of $\text{SNR}$, we can express it in terms of the batch-wise variance of $\nabla_{\theta}{\mathcal{L}_{\mathcal{B}}}$ as:

\[
 \nabla_{\theta}{\mathcal{L}_{\mathcal{B}}} - \mathbb{E}[\nabla_{\theta}{\mathcal{L}_{\mathcal{B}}}] = \epsilon_{\mathcal{B}}
\]

\begin{equation}
Var_{\mathcal{B}}(\nabla_{\theta}{\mathcal{L}_{\mathcal{B}}}) = \mathbb{E}_{\mathcal{B}}[\epsilon_{\mathcal{B}}^2].
\label{p3:f}
\end{equation}
\begin{equation}
\text{SNR} = \frac{\lVert \mathbb{E}_{\mathcal{B}}[\nabla_{\theta}{\mathcal{L}_{\mathcal{B}}}] \rVert_{2} }{ \lVert \epsilon_{\mathcal{B},rms} \rVert_{2}},
\label{p3:h}
\end{equation}

\noindent where $\epsilon_{\mathcal{B},rms}$ is the root mean square of the batch deviations $\epsilon_{\mathcal{B}}$. Therefore, from Eq. \ref{p3:h} when $\text{SNR} < 1$, the magnitude of the typical stochastic component $\lVert \epsilon_{rms} \rVert_{2}$ is greater than the magnitude of the deterministic component $\lVert \mathbb{E}[\nabla_{\theta}{\mathcal{L}_{\mathcal{B}}}] \rVert_{2}$ towards the update of $\theta$. This implies that the gradient noise will have a more pronounced effect than the signal in the direction of the parameter updates. Especially for cases where $\text{SNR} \ll 1$, the erratic behavior of the gradient due to the dominating noise among samples can be a characteristic of highly non-convex regions, caused by sample-disagreement, where the optimizer is unable to make meaningful progress. Similarly, when $\text{SNR} \gg 1$, there is a high certainty among samples for the next optimizer step. While this might be desirable in convex problems, this behavior can trap the optimization process in poor local minima from the early training stages for a non-convex loss landscape. Note that the stochasticity of SGD refers to the random selection of mini-batches. Although full-batch Adam is deterministic in that sense, we suggest that one can still measure the sample-wise stochasticity of each step based on the batch gradient noise $\sigma$, which is reflected in the SNR.

\subsection{Adam and step-wise SNR}

Adaptive-learning optimizers use $\text{SNR}$-related metrics to correct the next step for each parameter. Specifically for Adam \cite{kingma2014adam}, at each iteration $t$, the parameters are updated with the following rule:
 
\begin{equation}
\theta^{t+1} = \theta^t - \eta \cdot \mathcal{C},
\label{p3:a1}
\end{equation}

\noindent where $\Delta\theta = \eta\cdot\mathcal{C}$ is the parameter step and $\mathcal{C}$ the learning rate correction matrix given by:

\begin{equation}
\mathcal{C} = \frac{\hat{m}}{\sqrt{\hat{v}} + \epsilon} = \frac{ \mathbb{E}_{\beta_1}[\nabla_{\theta}{\mathcal{L}}]\sqrt{1-\beta_2^t} }{ \sqrt{\mathbb{E}_{\beta_2}[(\nabla_{\theta}{\mathcal{L}})^2]}(1-\beta_1^t) + \epsilon}, \ \ \forall \theta \in \Theta
\label{p3:a2}
\end{equation}

\noindent where $\hat{m}$ is the first moment and $\hat{v}$ is the second moment of the full-batch gradients, and $\mathbb{E}_{\beta}$ is the exponential moving average (EMA) of past steps. For all iterations past the initial training stage, the effect of the corrective terms is negligible with $\sqrt{1-\beta_2^t} \approx (1-\beta_1^t) \approx 1$. Thus, for $\epsilon \ll \sqrt{\mathbb{E}_{\beta_2}[(\nabla_{\theta}{\mathcal{L}})^2]}$, Eq. \ref{p3:a2} can be simplified as:

\begin{equation}
\mathcal{C} = \frac{ \mathbb{E}_{\beta_1}[\nabla_{\theta}{\mathcal{L}}] }{ \sqrt{\mathbb{E}_{\beta_2}[(\nabla_{\theta}{\mathcal{L}})^2]}}, \ \ \forall \theta \in \Theta.
\label{p3:a3}
\end{equation}

\begin{definition} 
To assess the dynamics of Adam during training, from Eq. \ref{p3:a3} we define $\text{SNR}_{Adam}$ as:

\begin{equation}
 \text{SNR}_{Adam} = \frac{\lVert \mathbb{E}_{\beta_1}[\nabla_{\theta}{\mathcal{L}}] \rVert_2}{\lVert \sqrt{\mathbb{E}_{\beta_2}[(\nabla_{\theta}{\mathcal{L}})^2]} \rVert_2},
\label{p3:a3.2}
\end{equation}
\noindent where the term $\lVert \mathbb{E}_{\beta_1}[\nabla_{\theta}{\mathcal{L}}] \rVert_2$ is the expected gradient (signal) and $\lVert \sqrt{\mathbb{E}_{\beta_2}[(\nabla_{\theta}{\mathcal{L}})^2]} \rVert_2$ is the $rms$ of the gradient (noise). Hence, $\text{SNR}_{Adam}$ represents the un-centered $\text{SNR}$ for each parameter $\theta$, with respect to the optimizer steps $t$ (resembling Eq. \ref{p3:g}).
\end{definition} 

Suppose that for an interval of $T$ iterations, the optimal control of the learning rate correction for each $\theta$ is achieved, then $\mathcal{C}$ will be at an equilibrium $\mathcal{C}^*$ for which $\hat{m} \propto \sqrt{\hat{v}}$ and:

\begin{equation}
\mathcal{C}^* \approx const \Rightarrow \text{SNR}_{Adam}^* \approx const, \ \ \forall t \in T.
\label{p3:a4}
\end{equation}

Therefore, if the optimal step is $\Delta \theta^* = \eta \cdot \mathcal{C}^*$, then for $\mathcal{C} < \mathcal{C}^*$ the current step $\Delta \theta$ decreases relative to the $\Delta \theta^*$ and vice versa. In simpler terms, in highly deterministic regions, the optimizer takes larger steps, while in highly stochastic regions, the optimizer is encouraged to take smaller steps, with a theoretical training equilibrium at $\text{SNR}_{Adam} = \text{SNR}_{Adam}^*$.

\subsection{Training dynamics and parameter sensitivity}
\label{p3:sensitivity}
Each step of the optimizer makes a parameter update that aims to better fit the outputs based on the inputs. However, the network parameters often have a different sensitivity to the loss, corresponding to scale variance of the landscape. To find a good minimum and exploit the full network architecture, the optimizer must respond to each parameter $\theta$ with similar sensitivity to fully capture their interdependencies during each backward pass. For example, if a parameter $\theta_1$ has a very sharp minimum which could restrict another parameter $\theta_2$ from progressing in that dimension, it requires a higher sensitivity of the optimizer, such that the different scales are comparable and meaningful progress can be made in all possible dimensions. Adam aims to do that by adjusting the learning rate correction $\mathcal{C}$ (sensitivity) based on each parameter's gradient response, leading to an adaptive scaling effect of the landscape for each dimension.

However, learning rate adjustments take into account the full-batch gradient which can result in prolonged training times, instabilities, or getting stuck in a sub-optimal space where not all the loss landscape trajectories have been explored. To illustrate this weakness, in Fig. \ref{p3:regimes}(c), we draw an indicative training path of a single parameter $\theta$ and three samples $x$, similar to \cite{domingos2020every}. At $t_1$, there is general agreement in the direction of samples. However, this path later splits into three possible branches: in branch A, the optimizer reaches a state where two (or more) samples cancel out. In branch B, the residual of sample $x_3$ remains unchanged during the recent steps, so for a quadratic loss, this means that $\nabla_{\theta}\mathcal{L}_{x_3} = 0$. Finally, branch C shows the path of a training equilibrium where the model is learning non-uniformly from different parts of $\mathcal{X}$. This process eventually convergences at a stationary solution with $\nabla_{\theta}\mathcal{L} = 0$, where the sample-wise gradients cancel out while being disproportional (as shown for step $t_{\infty}$), indicating that the model overfits to certain $x_i$ while underfitting to others. So while Adam aims to maintain a consistent direction for the full-batch trajectory ($\text{SNR}_{Adam} \approx const$), the individual sample-wise gradients can have a variable agreement between consecutive steps ($\text{SNR} \ll 1)$.

This can also be interpreted by the concept of sample typicality \cite{peng2019accelerating}, where the gradients of certain samples of a subset $\mathcal{H}$ provide the most informative search direction, while the remaining samples are grouped in subset $\mathcal{J}$. It has been shown that controlling the antagonism of $\mathcal{H}$ and $\mathcal{J}$ by assigning higher sampling probability to the first, accelerates the training convergence of SGD. Therefore, the subset $\mathcal{H}$ consists of samples that cancel out or that have become zero. Along this line, we can express the concept of general gradient typicality for a given sample set $\mathcal{X}$ when a size $m$ exists for which all equally sized subsets (or batches) are highly representative of the true gradient. This scenario implies an ideal gradient uniformity among samples, which is theoretically possible in the absence of gradient outliers and heteroscedastic noise in the labels $y_i$. 

\begin{definition}
A set $\mathcal{X}$ satisfies general gradient typicality when the gradients of all its subsets $\mathcal{H}_i$ with fixed size $m$ are approximately equal to the true gradient:

\begin{equation}
\frac{1}{m}\sum_{i \in \mathcal{H}_i} \nabla_{\theta}{\mathcal{L}_i} = \nabla_{\theta}{\mathcal{L}}, \ \ \forall \mathcal{H}_i \in \mathcal{X} 
\label{p3:t2}
\end{equation}
\end{definition} 

{To quantify the degree of typicality (or order) of the sample-wise gradients, we can also adopt the order parameter $\phi$, which is essentially a bounded form of $\text{SNR}$.}.

\begin{definition} 
{The order parameter $\phi$ measures the order of gradient alignment, capturing both directional and magnitude agreement of the gradients between any batches $\mathcal{B}$, and is given by:}

\begin{equation}
{\phi_{\nabla_{\mathcal{B}}\mathcal{L}} = \frac{\lVert \mathbb{E}[\nabla_{\theta}{\mathcal{L}_{\mathcal{B}}}] \rVert_2}{\lVert \sqrt{\mathbb{E}[(\nabla_{\theta}{\mathcal{L}_{\mathcal{B}}})^2]} \rVert_2}}.
\label{p3:phi}
\end{equation}
\end{definition} 

{When Eq. \ref{p3:t2} holds, then $\phi = 1$ and $\text{SNR} \rightarrow \infty$. However, given that batches are not identical, $\text{SNR}$ is generally bounded by the channel capacity at maximum compression \cite{shwartz2017opening}, so in practice $\phi < 1$ will always hold.}

\vspace{10pt}
\noindent \textbf{Proposition.} Starting from the expression $rms(x)^2 = \langle x \rangle^2 + std(x)^2$ for each network parameter, it can be shown that \text{SNR} is tightly linked to gradient order between samples through the following expression:
\begin{equation}
\text{SNR} = \frac{\phi}{\sqrt{1-\phi^2}},
\label{p3:t3}
\end{equation}

Moreover, from Eq. \ref{p3:t3} it can be shown that:

\begin{equation}
\text{SNR} > 1, \text{ for } \phi \in (\frac{\sqrt{2}}{2}, 1),
\label{p3:t4}
\end{equation}
\begin{equation}
\text{SNR} \approx \phi, \text{ for } \phi \ll 1.
\label{p3:t5}
\end{equation}

This means that when $\text{SNR}>1$ all batch-wise gradients become highly typical (Eq. \ref{p3:t4}), while for low gradient typicality, $\text{SNR}$ becomes approximately equal to $\phi$ (Eq. \ref{p3:t5}).

\subsection{Residual homogeneity}

As previously discussed, although Adam moderates each parameter step to leverage a consistent gradient signal, this does not necessarily guarantee that each sample $x_i$ is equally satisfied at the stationary solution (or at any point during training). In other words, the converged solution can have heterogeneous residuals $\mathcal{R}$.

We can analyze this statement through a simplified example. First we assume that a domain $\Omega$ has homogeneous residuals across the sample space if there exists a fixed volume $A$ for sub-domains $\Omega_i$ such that:
\begin{equation}
\int_{\Omega_a}|\mathcal{R}(x)|dx = \int_{\Omega_b}|\mathcal{R}(x)|dx, \forall \ \ \Omega_a,\Omega_b \in \Omega
\label{p3:e9.5}
\end{equation}
\[
\text{with } A(\mathcal{R}(x_i)) = const, \forall \ \ x_i \in \mathcal{X}
\]
Then, starting from the first-order Taylor expansion of the loss function $\mathcal{L}$, for two consecutive steps $t$ and $t+1$, we can make the following approximation:
\begin{equation}
\mathcal{L}(\theta^{t+1}) \approx \mathcal{L}(\theta^{t}) + (\theta^{t+1} - \theta^{t}) \nabla_{\theta}\mathcal{L}(\theta^{t}).
\label{p3:e10}
\end{equation}
Substituting the GD update rule in Eq. \ref{p3:e10}:
\begin{equation}
\mathcal{L}(\theta^{t+1}) - \mathcal{L}(\theta^{t}) \approx -\eta \cdot \lVert \nabla_{\theta}\mathcal{L}(\theta^{t}) \rVert_2^2
\label{p3:e11}
\end{equation}
At a stationary solution, $\mathcal{L}(\theta^{t+1}) - \mathcal{L}(\theta^{t}) = 0$ and:
\begin{equation}
\lVert \nabla_{\theta}\mathcal{L}(\theta^{t}) \rVert_2^2 = 0.
\label{p3:e12}
\end{equation}
By applying the linearity of differentiation for $p$ parameters and simplifying, we get the following:
\begin{equation}
\sum_{j=1}^p \left(\sum_{i=1}^n \mathcal{R}_i \frac{\partial \mathcal{R}_i}{\partial \theta_j}\right)^2 = 0 \Rightarrow \sum_{i=1}^n \mathcal{R}_i \frac{\partial \mathcal{R}_i}{\partial \theta} = 0, \forall \theta \in \Theta.
\label{p3:e15}
\end{equation}

From Eq. \ref{p3:e15} we see that although the terms $\mathcal{R}_i \frac{\partial \mathcal{R}_i}{\partial \theta_j}$ cancel out for all individual samples $x_i$, this condition is not enough to guarantee that the residual magnitudes $|\mathcal{R}_i|$ are homogeneous, or equal to 0 $\forall x_i \in d$. This can lead to a sub-optimal solution where Eq. \ref{p3:e9.5} is not satisfied, implying that different parts of the domain have converged while others have not. However, to reach a low test error (Eq. \ref{p3:l2}), residual homogeneity must also hold.

\subsection{Adaptive sample-wise weighting}
\label{p3:rba_weights}
A state where the sample gradients have equal sensitivity on the parameter updates can be achieved by enforcing uniform residuals for an interval of optimizer updates. If during that interval, $\mathcal{R}$ remains uniform, then $\nabla_{\theta}\mathcal{R}$ will also tend to be uniform. The residuals can still vary but at a similar rate. Therefore, the gradients $\nabla_{\theta}{\mathcal{L}}$ will also be uniform for any parameter $\theta$, since they consist of the product $\mathcal{R}\nabla_{\theta}\mathcal{R}$. Moreover, homogeneous residuals imply that information is propagating across the sample space evenly, which is a crucial aspect of training PINNs \cite{wang2024respecting}. In this work, the constraint of Eq. \ref{p3:e9.5} is softly enforced during training with a residual-based attention (RBA) re-weighting scheme. {This mechanism serves as a form of preconditioning by dynamically adjusting the contribution of individual sample gradients, based on the residual history from preceding steps. The proposed algorithm modifies the quadratic loss at each optimizer step as follows:}

\begin{equation}
\mathcal{L}(\theta, x) = \frac{1}{n}\sum_{i=1}^n \mathcal{R}_i(\theta, x_i)^2 \rightarrow \mathcal{L}^{\ast}(\theta, \lambda_i) = \frac{1}{n}\sum_{i=1}^n \mathcal(\lambda_i \cdot \mathcal{R}_i(\theta, x_i))^2,
\label{p3:e10.2}
\end{equation}

\noindent where $\mathcal{R}_i$ is the PDE residual at sample $i$ and $\lambda_i$ is its corresponding weight. Note that for simplicity, we consider a single-constraint optimization. For PINNs, this means that we focus on the residual loss while the boundary/initial condition losses are simply treated as added bias terms.

The weights $\lambda$ are calculated as a moving average of the relative residuals:

\begin{equation}
\lambda_i^{t+1} = \gamma\lambda_i^{t}+\eta^{\ast}\frac{|\mathcal{R}_i|}{\lVert \mathcal{R}_i \rVert_{\infty}}
\label{p3:e11f}
\end{equation}

Normalizing with $\lVert \mathcal{R}_i \rVert_{\infty}$ ensures that the maximum multiplier is bounded between [0, 1] when $\gamma = 1-\eta^{\ast}$. Thus, $\lambda_i$ is a weighting coefficient that rescales each term $\mathcal{R}_i \frac{\partial \mathcal{R}_i}{\partial \theta_j}$ with respect to their relative $|\mathcal{R}|$ history. Near convergence, for each parameter $j$ Eq. \ref{p3:e15} gives:

\begin{equation}
\sum_{i=1}^n \left(\lambda_i^2 \mathcal{R}_i \frac{\partial \mathcal{R}_i}{\partial \theta_j}\right) \rightarrow 0.
\label{p3:e12f}
\end{equation}

{In Eq. \ref{p3:e12f}, the sample-wise gradients are rescaled to emphasize unresolved outliers with persistently large residuals in the subsequent optimizer step, reducing the risk of converging to poor local minima. Intuitively, boosting the gradient (step) of larger residuals eventually leads to better residual homogeneity, decreasing the gradient variance. To analyze the behavior near a stationary solution where $\nabla_{\theta}\mathcal{L}(\theta, x) \rightarrow 0$, given a sufficiently small learning rate, we examine all possible cases that may arise:}

\vspace{10pt}
{\noindent\textbf{Case A:}
$\left|\frac{\partial \mathcal{R}_i}{\partial \theta}\right| > 0$ $\forall x_i \in \mathcal{X}$.
We split this into two subcases:}
\begin{itemize}
    \item[(i)] {All $\left|\mathcal{R}_i\right|\to 0$. \\
    Here, GD resolves every sample's residual; once all $\left|\mathcal{R}_i\right|$ vanish, 
    GD terminates in a solution with $\mathcal{L}(\theta, x) \approx 0$}.

    \item[(ii)] There exists {$x_i \in \mathcal{X}$ such that $\left|\mathcal{R}_i\right| > 0$. \\
    Because $\left|\frac{\partial \mathcal{R}_i}{\partial \theta}\right| > 0$, the gradient remains nonzero, 
    ensuring GD continues to update $\theta$ in a way that reduces $\left|\mathcal{R}_i\right|$}.
\end{itemize}

{\noindent\textbf{Case B:}
$\left|\frac{\partial \mathcal{R}_i}{\partial \theta}\right|\to 0$ for all $x_i \in \mathcal{X}$.
Again, we consider two subcases:}
\begin{itemize}
    \item[(i)] {All $\left|\mathcal{R}_i\right|\to 0$. \\
    This scenario is benign: if all residuals vanish along with their gradients, GD converges successfully}.

    \item[(ii)] There exists {$x_i \in \mathcal{X}$ such that $\left|\mathcal{R}_i\right| > 0$. \\
    Such a sample's residual cannot be reduced further by GD alone, because its (nearly) zero gradient 
    fails to direct any update. This leads to a suboptimal solution in which certain residuals remain nonzero}.
\end{itemize}

{The RBA algorithm specifically targets samples in case B(ii), where $|\mathcal{R}_i|$ is large but $|\frac{\partial \mathcal{R}_i}{\partial \theta}|$ is small. By boosting their weights, RBA prevents large-residual samples from stalling at near-zero gradients. Conversely, samples satisfying A(ii) (i.e., large $|\mathcal{R}_i |$ and large 
$|\frac{\partial \mathcal{R}_i}{\partial \theta}|$) tend to converge anyway, since their significant gradients drive them rapidly down the loss landscape. Consequently, their corresponding weights $\lambda_i$ diminish rapidly in RBA's moving-average mechanism. However, in special situations of case A(ii) (exceptionally stiff problems), where different samples' gradients cancel out (see Fig. \ref{p3:regimes}(c)), the optimizer may still settle in a suboptimal local minimum despite the re-weighting.}

\begin{figure}[H]
 \centering
 \includegraphics[width=.75\textwidth]{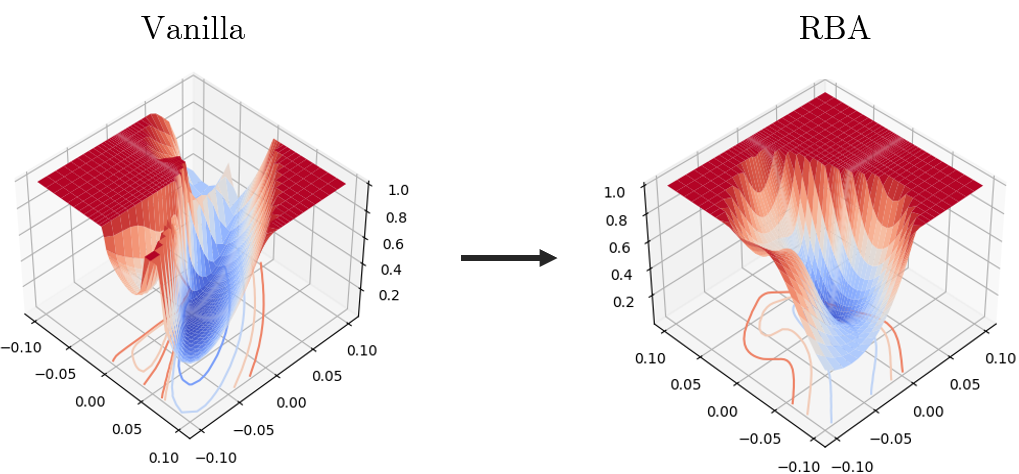}
 \caption[Vanilla vs RBA loss landscape]{{The RBA weights reduce the non-convexities caused by outliers, which lead to high gradient disagreement. While the vanilla loss landscape for Allen-Cahn is highly non-convex around the solution, the landscape of RBA appears significantly more convex.}}
 \label{p3:landscapes}
\end{figure}

{To demonstrate the effect of RBA, in Figure \ref{p3:landscapes} we vizualize the loss landscape around the solution for the vanilla and RBA Allen-Cahn case, based on the methodology proposed by \cite{li2018visualizing}. As the weights target the problematic samples of large residuals and small gradients, they achieves better gradient alignment, which makes the optimizer equally sensitive to each sample.  Thus, reducing gradient variance eventually eliminates the non-convexities caused by unresolved outliers. We further note that this method is not PINN-specific, as it can be applied to any quadratic loss function, which exhibits this type of slow-converging outliers leading to increased gradient variance and instabilities.}

\begin{definition} 
{The residual homogeneity can also be measured using the corresponding order parameter:}
\begin{equation}
{\phi_{RBA} = \frac{ \mathbb{E}[\lambda_{\mathcal{B}}]}{\sqrt{\mathbb{E}[\lambda_{\mathcal{B}}^2}]}}.
\label{p3:phi_rba}
\end{equation}
\end{definition}

\section{Results}

\subsection{DE and convergence}

\begin{figure}[H]
 \centering
 \includegraphics[width=0.9\textwidth]{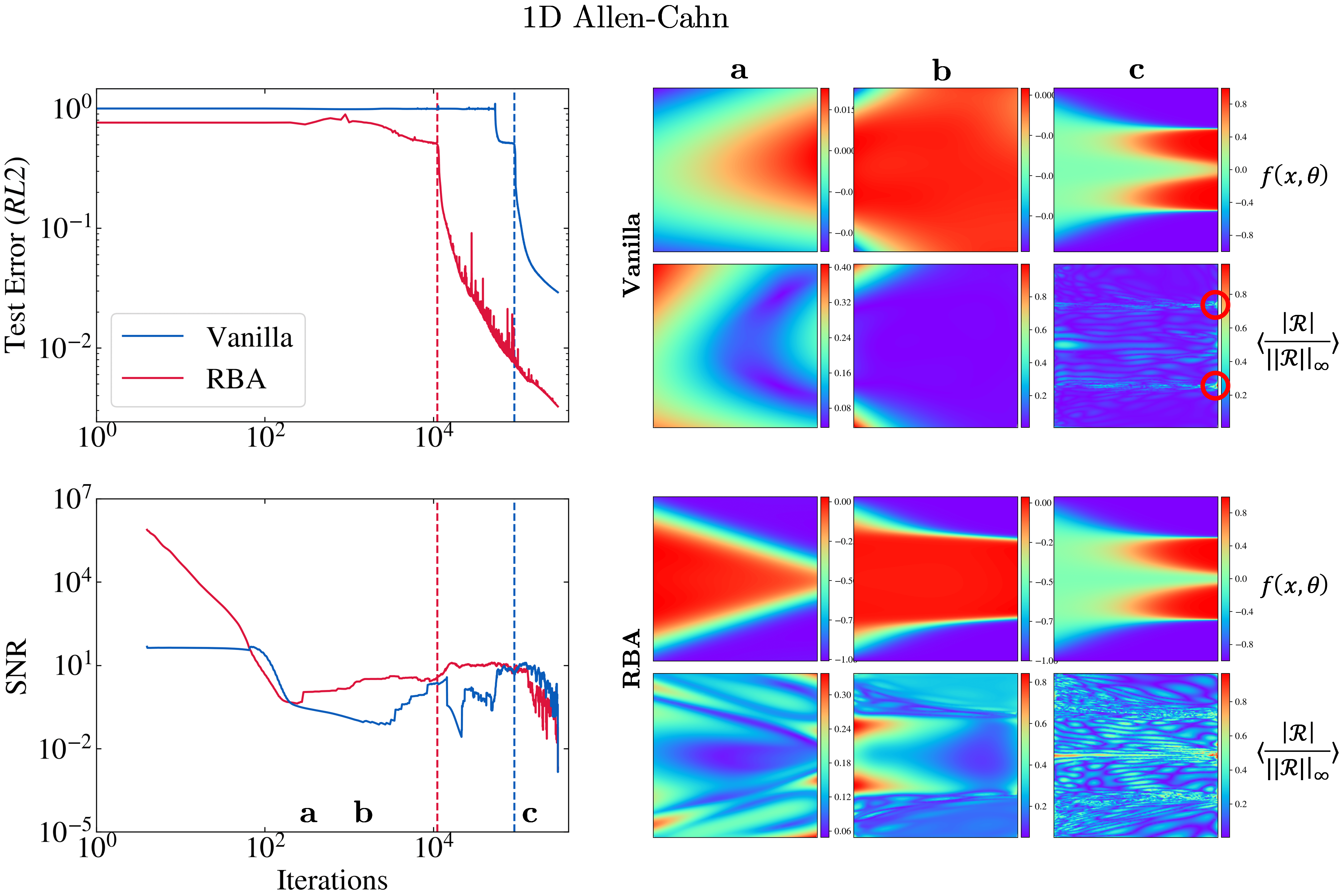}
 \caption[Allen-Cahn training dynamics]{ \textbf{Allen-Cahn training dynamics -} \textbf{Fitting}: Both vanilla and RBA experience a rapid decrease in their \text{SNR} prior to diffusion, while the test error is still stagnant. \textbf{Diffusion}: Both models enter diffusion as the \text{SNR} becomes more stochastic. The vanilla model does not converge, and the residuals $\mathcal{R}$ are not homogeneous, while RBA slowly converges with a characteristic double-jet shape for the residuals (snapshots (a,b)). \textbf{DE} (dashed lines): The RBA model enters the equilibrium phase first, while $\mathcal{R}$ is highly homogeneous, accompanied by a steep test error convergence. The vanilla $\text{SNR}$ shows a small step increase towards DE, and its RL2 decreases, but the residuals are still heterogeneous (snapshot (b)). On the other hand, the RBA model residuals appear highly diffused, accompanied by a steep RL2 convergence. At snapshot (c), the vanilla finally converges to a sub-optimal solution, where $\mathcal{R}$ appears more diffused than before but still has a few samples which have not converged (red circles), causing a less-pronounced color spectrum of $\mathcal{R}$. The RBA method keeps decreasing the test error further, while $\mathcal{R}$ is highly homogeneous, implying that different regions of $\mathcal{X}$ are equally satisfied. \textbf{Accuracy}: For a 10\% test error, vanilla takes 100k iterations while RBA takes 10k, thus RBA is 10 times faster than vanilla.}
 \label{p3:res_AC}
\end{figure}

\begin{figure}[H]
 \centering
 \includegraphics[width=1\textwidth]{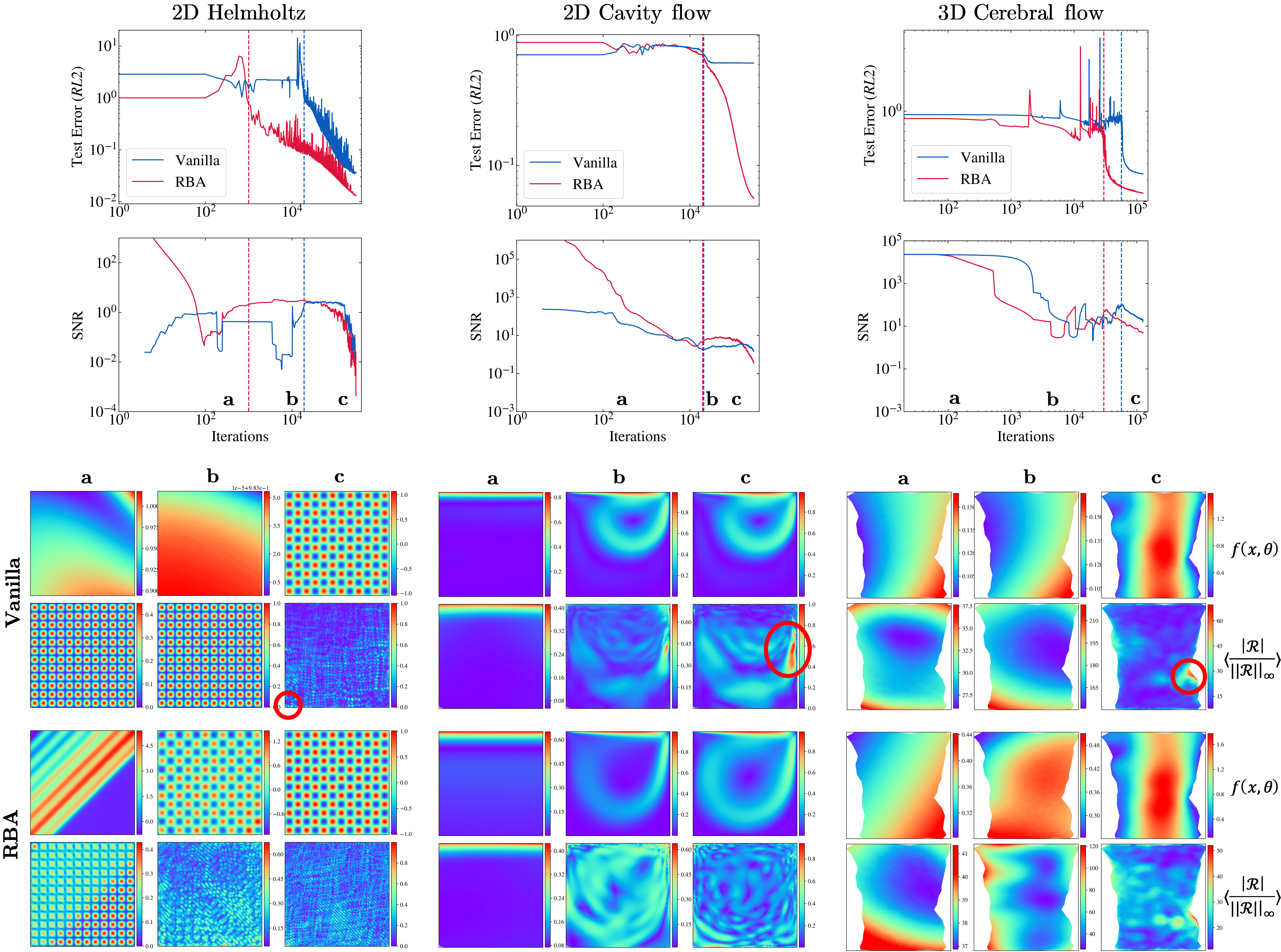}
 \caption[Training dynamics]{\textbf{Training dynamics -} \textbf{Fitting}: Both vanilla and RBA models undergo the usual rapid transition from high to low \text{SNR}, where the optimizer starts reacting to the backpropagated gradients until each batch starts differentiating from each other. During this phase, the test error does not reduce. \textbf{Diffusion:} For all cases, the error starts converging slowly and the \text{SNR} further decreases, while the residuals are spreading across the domain. Specifically for Helmholtz, the \text{SNR} of both models is more stochastic, but the vanilla $\text{SNR}$ further decreases while the $\text{SNR}$ of RBA remains closer to DE. While the residuals of the vanilla are not homogeneous, they start diffusing for RBA (snapshots (a,b)). \textbf{DE} (dashed lines): For all cases, the sudden transition into DE is met by a sudden increase of the \text{SNR} to a stable equilibrium and a simultaneous drop in the test error. The vanilla $\mathcal{R}$ is still heterogeneous, but for RBA, it appears more homogeneous (snapshot (b)). For Helmholtz, when RBA goes into DE, marking the main convergence of the test error, the vanilla $\text{SNR}$ dives deeper into the stochastic regime where the test error cannot be decreased. At snapshot (c), all three cases are now in equilibrium with corresponding test error convergence. However, the residuals of the vanilla model appear less diffused compared to the RBA, which results in a suboptimal solution. The RBA method keeps decreasing the test error further, while $\mathcal{R}$ is highly homogeneous, implying that different regions of $\mathcal{X}$ are equally satisfied. Furthermore, for cavity and cerebral flows the vanilla model converges to a stationary solution with high test error due to the heterogeneous residual regions (circled in red), which hinder the information flow to the rest of the domain. Since the region with accumulated high residuals cannot be resolved, the vanilla cannot generalize well, especially for the cavity flow, where it gets stuck around 60\% test error. The RBA model, however, distributes the residuals evenly across the cavity and generalizes with a final test error of about 6\%. Note how the \text{SNR} remains above $\mathcal{O}(1)$, which is reflected by higher agreement/determinism of the steps and smoother convergence.}
 \label{p3:new_res}
\end{figure}

To analyze the phase transitions in PINN learning, we train three standard PINN benchmarks: Allen-Cahn (Fig. \ref{p3:res_AC}), Helmholtz and lid-driven cavity flow (Re=1000) (Fig. \ref{p3:new_res}). In addition, a real-world inverse problem is modelled by solving the transient 3D Navier-Stokes equations to infer the cerebrospinal fluid (CSF) flow (Fig. \ref{p3:new_res}) within a murine brain. Both Figs. \ref{p3:res_AC}, \ref{p3:new_res} show the test error convergence (test error), the $\text{SNR}$ curves, and three snapshots (a-c) of the prediction $f(x, \theta)$, along with the RBA weights across the solution domain. We always compare the RBA with the vanilla model and plot the RBA weights for both, even though they are not used in the optimization process of the vanilla. The colored dashed lines mark the DE phase transition for each model. To make the sample-wise gradient computation more efficient, we split the collocation points into batches and treat the initial and boundary conditions as constant terms added to each batch. For the 3D transient flow case we visualize a slice of the 3D geometry and average the residuals over 3 equidistant snapshots of the cardiac pulse. {For this study we fix the number of layers to 6 with 128 neurons each, following recent studies in PINNs which show good performance on similar problems \cite{anagnostopoulos2024residual,zhang2023dasa,wang2024respecting}. All networks are trained with Adam with initial learning rate of 0.001 and tanh activations. We note that we don’t perform any parametric investigation regarding the architecture, but future studies will aim to look into the combinations of various architectures and hyperparameters, in order quantify how it affects the training dynamics and SNR phases.} The SNR plots for all cases have been smoothed using a simple moving average (SMA) of 100 past iterations. In the Appendix we include additional SNR results and further implementation details for each case. The relevant code is available on github.com/soanagno/diffusion-equilibrium.

As evident from these results, it is after the diffusion process has reached its equilibrium when the models experience the steepest convergence of the test error. Moreover, the RBA models induce higher homogeneity of the residuals, resulting in better generalization. At the phase transition, the \text{SNR} increases towards the deterministic regime where it stabilizes {($\frac{\partial}{\partial t}\text{SNR}\approx 0$), as the gradients agree on the direction of the optimizer. Interestingly, most cases converge non-monotonically after DE, which could be related to the ``edge of stability" phase, where the inherent self-stabilization mechanism of GD keeps the system in a stable equilibrium when the Hessian sharpness increases higher than the critical value \cite{damian2022self}}. On the other hand, the convergence of the cavity flow appears monotonic (Fig. \ref{p3:new_res}), which could be attributed to the $\text{SNR}$ being $ > \mathcal{O}(1)$. Furthermore, both vanilla and RBA models for the cerebral flow case exhibit multiple SNR jumps prior to the main equilibrium phase, which always coincide with partial test error decreases (but do not succeed in overcoming the local minima). Finally, it appears that the DE phase for the cerebral case is very brief, which causes the $\text{SNR}$ to immediately decrease after transition since Adam has already converged.

These results align with the intuition that training in the diffusion phase with residual homogeneity is essential, especially in the context of PINNs, where individual samples are explicitly interdependent based on the governing PDEs. Additionally, in all four cases the DE phase is accompanied by highly homogeneous residuals for the RBA model, while the vanilla model has regions of accumulated residuals {(circled in red Fig. \ref{p3:new_res})}, which lead to sub-optimal solution accuracy. {However, Adam converges even though these high-residual samples don't get minimized, which implies that their corresponding neural gradients are zero or cancel out, forming a local minimum, as discussed in Sections \ref{p3:sensitivity}-\ref{p3:rba_weights}.} Since the RBA models resolve these regions, this results in about one order of magnitude faster convergence.

Another interesting observation from these results is that the onset of DE and the SNR stabilization region appears to have case-specific bounds, which is likely related to the characteristics of the loss landscape for each problem (e.g. the sharpness of the Hessian). Although not within the scope of this study, determining the existence of upper bounds or critical regions for the SNR, as well as estimates for the DE transition, would be of great value for the decision-making during the model development stage. For example, knowing such characteristics regarding the training dynamics a priori could provide more quantitative methods for the selection of the network architecture based on the problem description, the specific PDE, boundary condition terms, and the quality of the data for-data driven problems.

\subsection{{Importance of gradient and residual order}}

\begin{figure}[H]
 \centering
 \includegraphics[width=.95\textwidth]{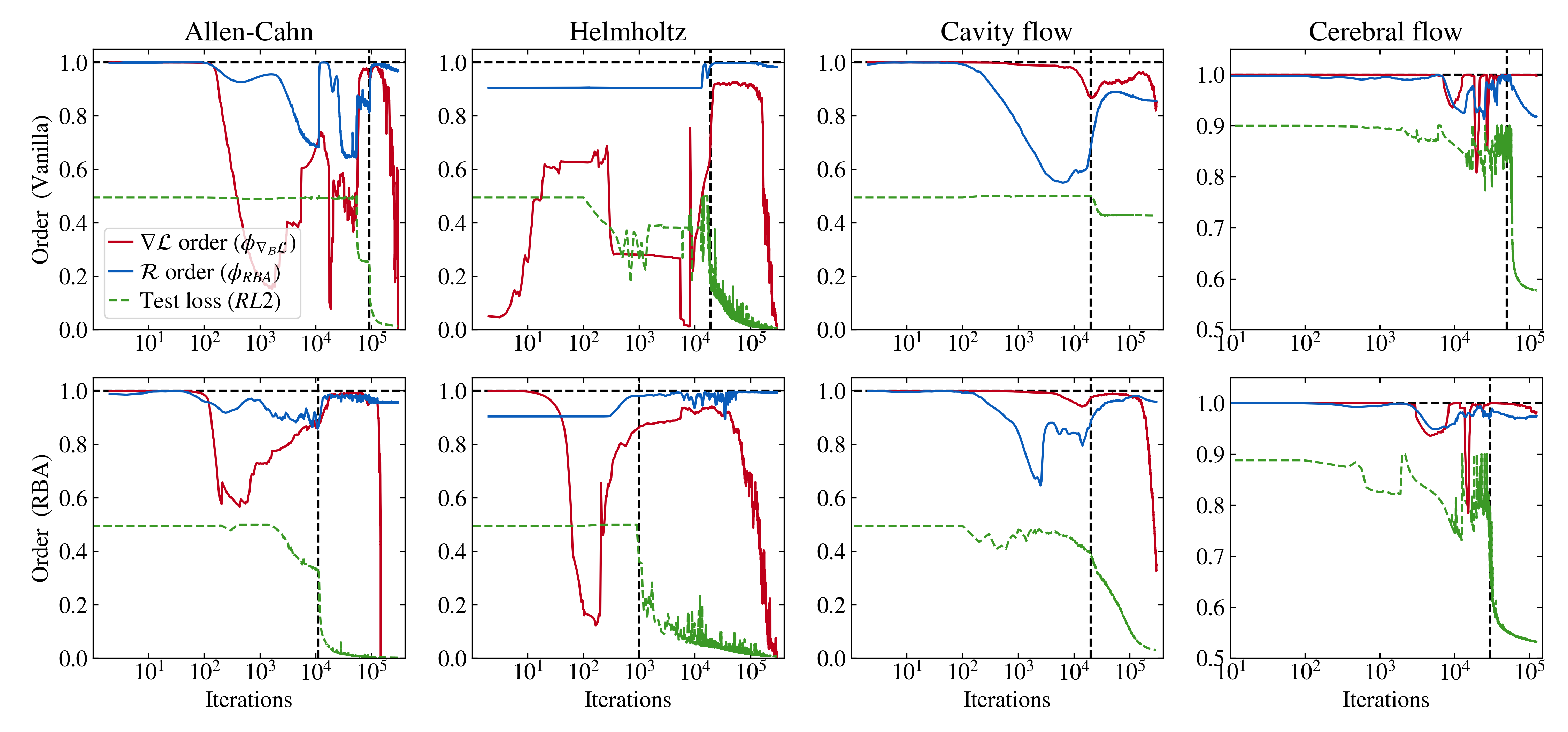}
 \caption[Gradient vs residual order]{ {\textbf{Gradient vs residual order}: The significance of residual homogeneity is demonstrated by plotting the residual order $\phi_{RBA}$ and the order of gradients $\phi_{\nabla_{\mathcal{B}\mathcal{L}}}$, alongside the test error (RL2). When high gradient order is met with high residual order, then the optimizer can generalize best. The residuals may become homogeneous before DE (see vanilla Allen-Cahn at 10k), keeping optimization stagnant, but when their order is combined with high gradient order (both $\phi\approx 1$), the onset of DE leads to stable convergence. In general, higher gradient and residual order correlates with better generalization at the phase transition of DE (vertical dashed lines).}}
 \label{p3:res_phi}
\end{figure}

{The importance of residual homogeneity is shown in Fig. \ref{p3:res_phi}, where we plot the order parameter of the gradients ($\phi_{\nabla_{\mathcal{B}}\mathcal{L}}$) and residuals ($\phi_{RBA}$), alongside the test error (RL2). As the order parameter is bounded between 0 and 1 this helps us make meaningful comparison between gradient alignment and residual homogeneity. Evident from all cases is that gradient order is not enough for convergence, but when is met with high residual order, the optimizer can stably converge. Note that the residuals might become homogeneous prior to the gradient alignment (phase transition), but this will still keep the optimization stagnant. As an example, the vanilla Allen-Cahn residuals become homogeneous around 10k iterations but since $\phi_{\nabla_{\mathcal{B}}\mathcal{L}}$ is low, there is no convergence. Around 60k iterations there is a second $\phi_{RBA}$ increase which does not reach the maximum order of 1, but is coupled with an increase in $\phi_{\nabla_{\mathcal{B}}\mathcal{L}}$, leading to convergence to a local minimum. At 100k iterations $\phi_{RBA}$ increases again, this time close to 1, which causes an even lower RL2 convergence. On the other hand, the RBA Allen-Cahn model maintains the residual homogeneity after 10k iterations along with high gradient order, leading to better generalization. In general, a common pattern is observed for all cases: higher gradient and residual order leads to better generalization.}

\begin{figure}[H]
 \centering
 \includegraphics[width=.95\textwidth]{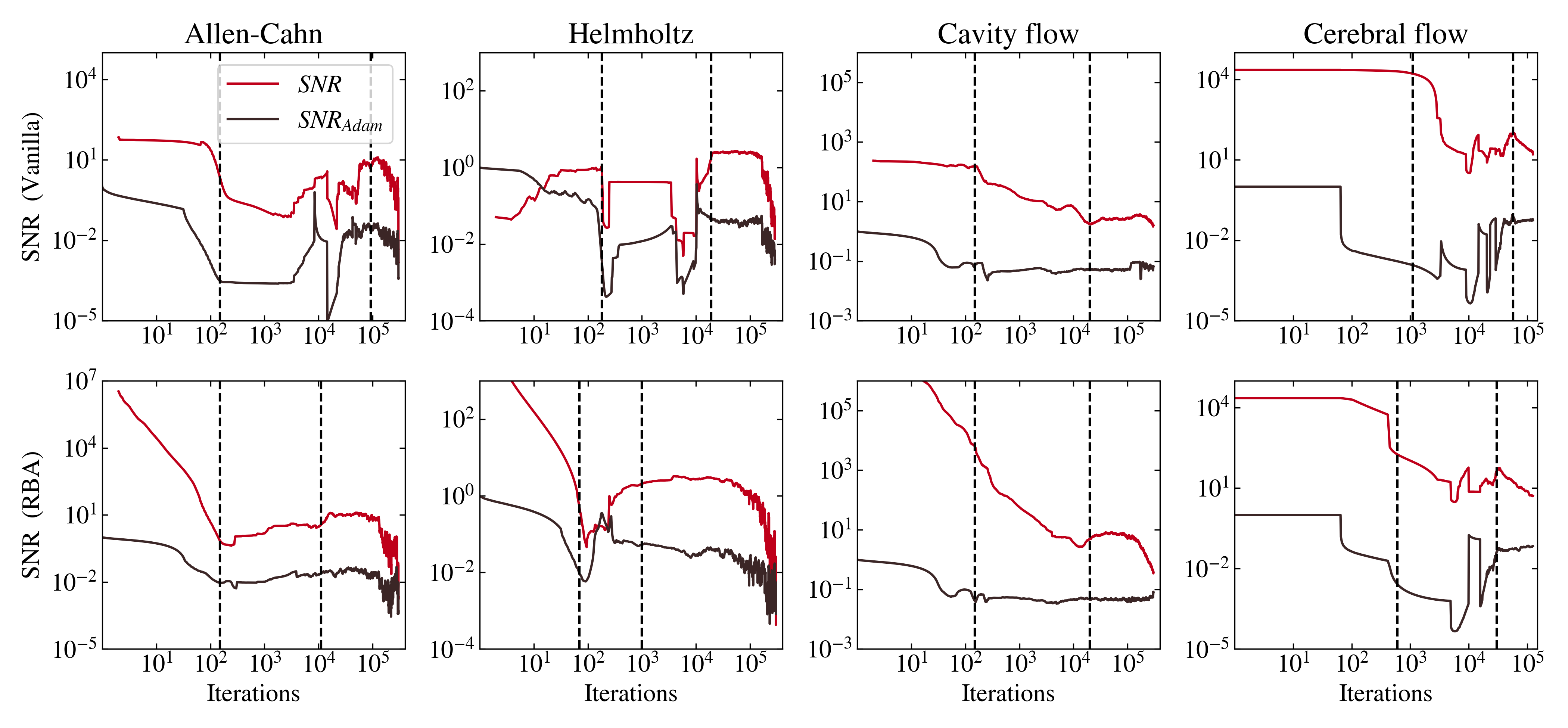}
 \caption[Correlation of $\text{SNR}$ and $\text{SNR}_{Adam}$]{ \textbf{Correlation of $\text{SNR}$ and $\text{SNR}_{Adam}$:} For all cases there is a high correlation of \text{SNR} with $\text{SNR}_{Adam}$. Namely for all RBA models, $\text{SNR}$ closely follows the curve of $\text{SNR}_{Adam}$, while at DE, they both reach a stable equilibrium which coincides with the steepest test error convergence. The two vertical dashed lines are the``fitting/diffusion" and ``diffusion/DE" phase transitions. {This correlation implies that SNR is driven by Adam, which eventually reduces the stochasticity among samples, and stabilizes convergence. This also indicates that an appropriate step for each network parameter has been found such that every $p$ is equally sensitive to each training sample $x_i$.} Notably, although $\text{SNR}$ undergoes a high-to-low transition, the optimal convergence is met at DE, when the gradients become homogeneous.}
 \label{p3:adam_rms}
\end{figure}

Adam adapts the learning rate of each parameter through the iteration-wise gradient SNR, which we define as $\text{SNR}_{Adam}$. An important observation we made for all cases of this study is the correlation of SNR with $\text{SNR}_{Adam}$, as shown in Figure \ref{p3:adam_rms}. More specifically, for all RBA models $\text{SNR}_{Adam}$ closely follows the curve of $\text{SNR}$, while at the phase transition, they both reach a stable equilibrium which coincides with the steepest test error convergence. Note that the two vertical dashed lines in Fig. \ref{p3:adam_rms} correspond to ``fitting/diffusion" and ``diffusion/DE" phase transitions. Therefore, it appears that Adam is able to converge more optimally when the $\text{SNR}$ is at an equilibrium with homogeneous gradients. {This correlation implies that SNR is driven by Adam, which eventually reduces the stochasticity among samples, and stabilizes convergence. We also see that for all cases Adam achieves an equilibrium where $\frac{\partial}{\partial t}SNR_{Adam}\approx0$, indicating that an appropriate step for each network parameter has been found such that every $p$ is equally sensitive to each training sample $x_i$}. Notably, although $\text{SNR}$ undergoes a high-to-low transition, the optimal convergence is met during DE. During that phase, the abrupt increase in the $\text{SNR}$ remains several orders of magnitude lower than the fitting phase $\text{SNR}$. 

\subsection{Information compression}

In IB theory, information compression serves as the primary explanation for the phase transition of the gradients, which occurs when the layers' activations saturate at the bounds of the activation function, given that such bounds exist. During this phase, the network maximizes the information of the layer activations $\mathcal{T}$ with respect to the output $I(\mathcal{T};\mathcal{Y})$, while minimizing (compressing) their information based on the input $I(\mathcal{X}; \mathcal{T})$. However, this claim has been debated through several studies. For instance, the phase transition has been replicated even with ReLU activation, which is not bounded like tanh \cite{saxe2019information}. Another important observation is that for a continuous range the information is infinite, so the measured information can be quite sensitive to the discretization method which is usually adopted \cite{lorenzen2021information,geiger2021information}. Thus, to interpret how the different layers carry the information from the input to the output, the range of the activations must be carefully discretized prior to applying the appropriate strategy of quantifying the information \cite{goldfeld2018estimating}. However, a universal method that yields the exact quantification of $I$ remains an open question. 

Herein, we investigate the compression phenomenon by focusing on $I(\mathcal{X}; T)$ and its behavior during the phase transition. The mutual information, under the assumption that $\mathcal{T}$ is a deterministic function of $\mathcal{X}$, is given by:

\begin{equation}
I(\mathcal{X};\mathcal{T}) = H(\mathcal{T}) - H(\mathcal{T}|\mathcal{X}) = H(\mathcal{T}) = \sum_{i=1}^{i=N}p_ilog_2(p_i).
\label{p3:i1}
\end{equation}

To discretize the activations for each layer, we calculate the mean range of each layer's neurons and create 30 equidistant bins within that range. This relative-binning approach has the advantage of considering the relative range of each layer (similar to \cite{basirat2021geometric}), which might vary depending on the network properties and stage of training (e.g. layers with smaller range get appropriate granularity). Therefore, each neuron takes 30 possible values, and each layer $l$ represents a string of $p_l$ digits (representation), where $p$ are the parameters of the layer. If each unique input $x_i$ is correlated to a unique representation of the layer, we assume that the layer carries the maximum information possible to distinguish the inputs, equal to $log_2(n)$ bits. The entropy of Eq. \ref{p3:i1} is finally calculated using the probability of occurrence of each unique representation $\mathcal{T}$ for each input $\mathcal{X}$.

\begin{figure}[H]
 \centering
 \includegraphics[width=1.0\textwidth]{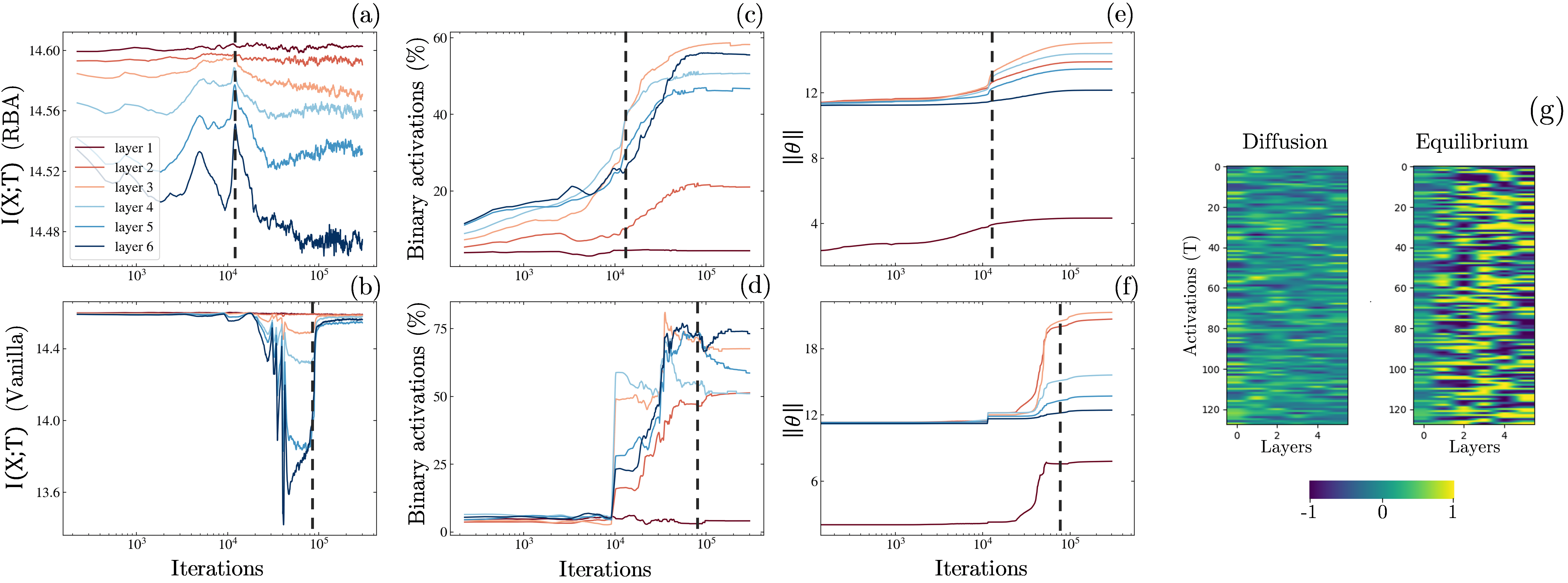}
 \caption[Information compression]{\textbf{(a-f) Compressed representation and phase transition for Allen-Cahn:} $I(\mathcal{X}; \mathcal{T})$ curves for each layer in bits (a, b), percentage of binary activations per layer (c, d) and parameter norm (e, f). Due to the discretization method, there is only a negligible $I$ decrease after the DE for RBA. This implies that even the deeper layers respond to the inputs during training, although with a smaller range for the first iterations (a). For vanilla, there is a more considerable information loss at the onset of diffusion, which get recovered after DE (b). {Due to the gradient alignment (high SNR), the parameters $\theta$ rapidly increase at the DE phase transition (dashed lines), which causes the $tanh$ activation to saturate at its boundaries}, compressing the representation of the input information (e-f). Note that the middle layers show the highest compression, reminiscent of an encoding-decoding architecture. \textbf{(g) Saturation of deep layers:} Indicative activation values for a random input during the diffusion (left) and equilibrium (right) phase. As the weights increase, the tanh activation saturates at its boundaries, compressing the representation of the input information.}
 \label{p3:mi}
\end{figure}

Our results indicate that, although there is some lossy compression after the phase transition (decrease in $I(\mathcal{X}; \mathcal{T}$)), most of the information is available across all layers from the early iterations, implying that the network can be sensitive enough to pass the information to the deeper layers, given that $I(\mathcal{X};\mathcal{T})$ is calculated with the relative-binning (Fig. \ref{p3:mi}(a, b)). This information compression has been characterised as ``geometric” \cite{geiger2021information} or ``clustering” \cite{goldfeld2018estimating} due to the saturation of $\mathcal{T}$, while the minimum information for distinguishing each input is preserved. “Binary” activations which maintain the information essentially compress the representations by exploiting redundancies within the data, leading to a near-lossless ``information-theoretic” compression. We note that measuring all layers with the absolute binning (as some of the earlier IB studies) produced mixed results: during the fitting phase, the range of the deep layers was smaller than the layers close to the input, but after diffusion, the range of the deep layers was larger. Thus, since the first layers do not saturate, this leads to higher $I(\mathcal{X};\mathcal{T})$ in the deep layers during diffusion, which, based on the Markovian chain assumption is not consistent. However, we suggest that the information compression does not necessarily imply dramatic theoretic-information loss in the layers but mainly saturation of the activations. As shown in Fig. \ref{p3:mi}(c, d), the deeper layers become more saturated/binary as the training progresses, with a sudden increase during the equilibrium phase. In essence, it appears that the information required to distinguish most inputs is present from early on. Still, the network becomes more saturated and achieves a compressed representation during this stable phase, mainly for the deeper layers.

Since our results align with the idea that the quantization of $\mathcal{T}$ can reduce the compression effect when measuring $I(\mathcal{X};\mathcal{T})$, we also show the percentage of saturated/binarized activations in Fig. \ref{p3:mi}(c, d), which indicate that geometric compression is present and it is not necessarily linked to theoretic-information compression, as also proposed by \cite{geiger2021information}. During {DE phase, due to the aligment of the neural gradients of most layers, the saturation of the tanh activations maximizes}, indicating a compressed/efficient representation. This is also captured by the network parameter norm $\lVert \theta \rVert$, which shows that the middle layers compress more (Fig. \ref{p3:mi}(e, f)). This phenomenon is reminiscent of an encoding-decoding process and was also present for most other benchmark cases (not shown here). Only for Helmholtz, the layers were compressed with a hierarchical order (ascending $\lVert \theta \rVert$ from the first to the last layer). Moreover, although the vanilla model achieves a compressed representation (Fig. \ref{p3:mi}(f)), its layer activations undergo a less smooth transition (Fig. \ref{p3:mi}(d)).

Finally, the ``binarization'' of the layers is further demonstrated in Figure \ref{p3:mi}(g), where we show the activations for a random input prior to and after DE. The deeper layers achieve a binary representation of the input, which is more compressed than the first layers. Finally, an interesting observation is that geometric compression is not necessary for good generalization performance \cite{geiger2021information}, since the vanilla model (Fig. \ref{p3:mi}(d)) compresses more than RBA (Fig. \ref{p3:mi}(c)), but the latter generalizes better (Fig. \ref{p3:res_AC}).

\section{Summary}
\label{p3:Summary}

In this study, we attribute the phase transitions in PINNs to different degrees of gradient order, and show that {residual homogenity at the DE phase is crucial for the optimizer stability and convergence. Moreover, we propose a weighting scheme (RBA) to increase the residual order and onset this transition faster, improving generalization. We note that although this study focuses on PINNs, the theoretical basis of our observations could also apply to a broader range of ML problems. Our re-weighting scheme is also a generic algorithm with negligible computational cost and easy implementation, which could possibly enhance the performance of other ML problems with slow-converging outliers, apart from PINNs.} 

More specifically, we empirically demonstrate the existence of phase transitions proposed by the IB theory (fitting/diffusion) for PINNs trained with Adam. Additionally, we identify a previously undisclosed third phase, which we name ``diffusion equilibrium”, {an equilibrium phase of batch-wise gradient agreement associated with generalization performance.} Our findings indicate that PINNs achieve optimal learning when residuals are also diffused uniformly across the sample space, leading to increased stability of Adam's learning rates. To further support this hypothesis, we introduce a re-weighting technique (RBA) to improve the vanilla PINN performance, which induces residual homogeneity faster, and improves generalization.

Based on IB theory, the training undergoes a high-to-low SNR transition, and since diffusion is a stochastic process, it is accompanied by a low SNR (increased noise). While fitting is usually a fast, deterministic phase of large gradients with minimal (useful) information flow (i.e., low $I(\mathcal{T};\mathcal{Y})$), it is during the slow diffusion phase when most learning happens, as each batch starts ``reacting’’ to the optimizer steps, exploring the non-convex landscape due to the increased stochasticity. Although both IB models and our models start converging during diffusion, for PINNs it seems that the fastest convergence happens when the DE is reached, representing a state of high information flow. In addition, we observe that this transition is also tightly linked with the optimizer stability, even in the presence of non-monotonic convergence, which we attribute to the self-stabilizing regime of the “edge of stability”. This phase could potentially describe optimal information processing in fully-connected neural networks besides PINNs. {Nevertheless, it remains unclear why the onset of the DE phase, which leads to a rapid convergence, is a first-order transition and not a progressive process. More work needs to be conducted regarding the theoretical aspect of the SNR and the generality of DE, and more specifically on the existence of critical SNR regions depending on model characteristics. This could be further investigated by convergence analysis for PINNs and the study of analytical bounds related to SNR. Future studies will also aim to test RBA weights on a broader range of architectures (e.g., CNNs) and exclusively data-driven problems.}

Finally, we investigate the relationship between SNR phase transition and information compression, observing a compression phenomenon due to the saturation of the network activations. At the abrupt phase transition, the network parameters rapidly increase {due to the highly aligned gradients}, resulting in the saturation of neuron activations. Employing a relative-binning strategy, we find that while there is negligible information loss in deeper layers, there exists a hierarchy in the information content across layers, aligning with the principles of the IB theory. Our analysis reveals that middle layers exhibit the most pronounced saturation during this ``binarization" process, reminiscent of an encoder-decoder mechanism within the fully-connected architecture. However, a more compressed representation does not necessarily lead to higher information-theoretic loss or better generalization. We believe this novel connection between IB theory and scientific ML could open new possibilities for quantifying the performance of the many versions of PINNs and even neural operators, as it gives a consistent description of the training dynamics.

\section*{Acknowledgements}
SJA and NS thank the Swiss National Science Foundation grant ``Hemodynamics of Physiological Aging" (Grant nr $205321\_197234$). JDT and GEK acknowledge support by the NIH grant R01AT012312, the DOE SEA-CROGS project (DE-SC0023191), the MURI-AFOSR FA9550-20-1-0358 project, and the ONR Vannevar Bush Faculty Fellowship (N00014-22-1-2795).


\appendix

\section{Batch-size-independent SNR}

The batch-wise SNR of Eq. \ref{p3:g} is used to circumvent the computational cost of performing a backward pass for each sample. However, the sample-wise SNR would depend on the batch construction. When each batch consists of \(m\) independent sample gradients with mean \(\mu\) and total sample-level variance \(\nu^2\), the variance of the batch mean scales as \(\nu^2/m\). Hence, the batch-independent SNR ($\text{SNR}_S$) is given by:

\begin{equation}
    \mathrm{SNR}^2
    =m\,\frac{\|\mu\|_2^2}{\nu^2} \Rightarrow \text{SNR}_s
    =\frac{\text{SNR}}{\sqrt m}.
    \label{eq:snr_batch_size}
\end{equation}

\begin{figure}[H]
 \centering
 \includegraphics[width=.5\textwidth]{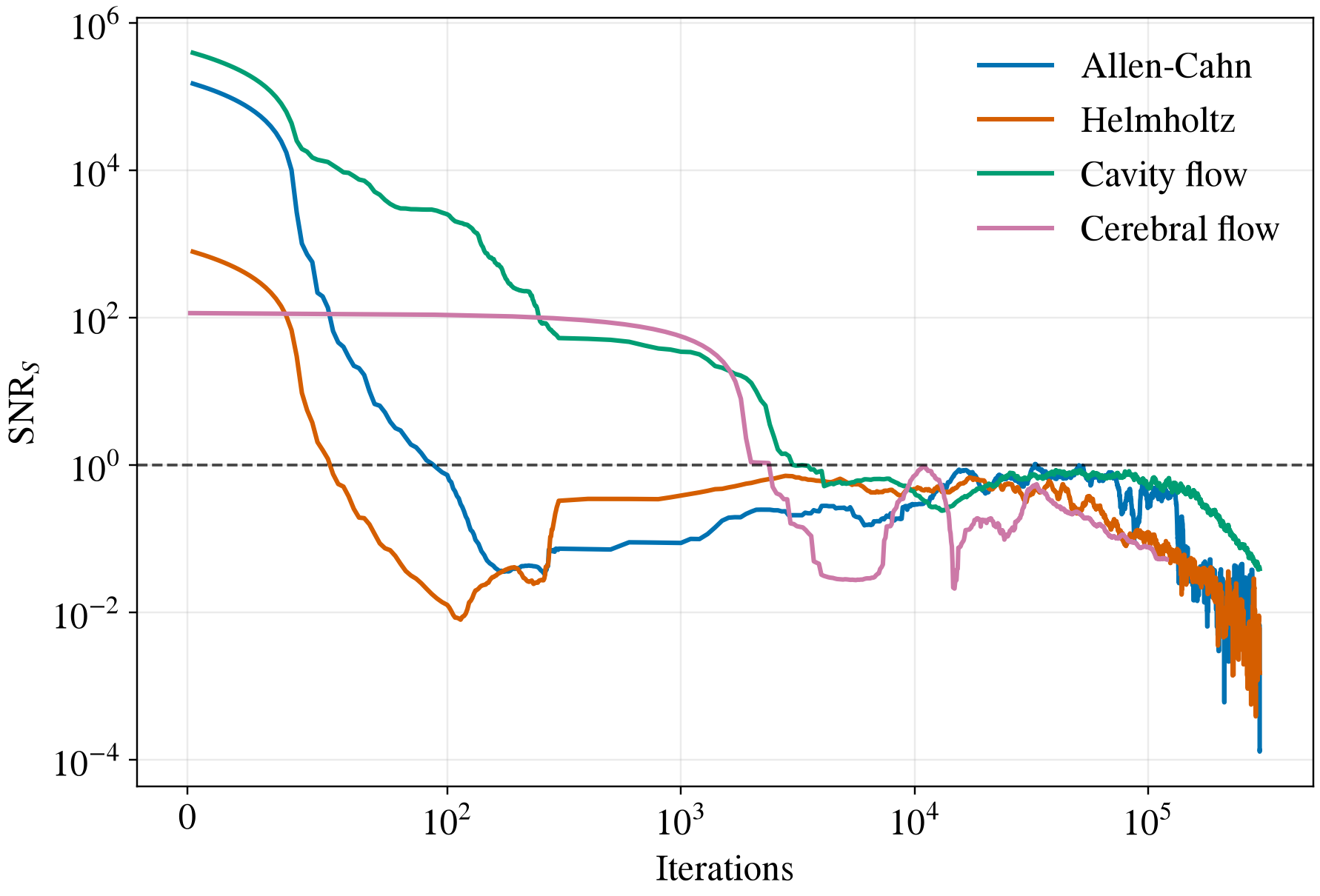}
 \caption[Layer-wise SNR]{ \textbf{Batch-size-independent SNR:} Curves of $\text{SNR}_S = \text{SNR}/\sqrt m$ for the four PINN problems. During the equilibrium/convergence phase, all cases approach an $\mathcal{O}(1)$ alignment level slightly below $\text{SNR}_S=1$, indicating comparable coherent (signal) and fluctuating (noise) sample-gradient contributions.}
 \label{p0:snr_s}
\end{figure}

Figure \ref{p0:snr_s} shows $\text{SNR}_S$ curves (batch-wise $\text{SNR}$ times a $1/\sqrt m$ factor), where all cases sit just below unity during the equilibrium phase. This indicates that the mean sample-gradient magnitude and the sample-to-sample gradient fluctuations become comparable in scale. The fact that the curves remain slightly below $\text{SNR}_S=1$ suggests a regime in which fluctuations are marginally dominant, while the gradient field is nevertheless substantially more aligned than in the preceding low-SNR phase. For Allen–Cahn and Helmholtz, we approximate the effective batch sizes $m_{eff} =m/2$ and $m_{eff} =m/16$, respectively, to account for the corresponding spatial symmetries of the solution.

\section{Proposition}

\begin{proof}
    
Adding \(\text{rms}(x_i)^2 = \text{mean}(x_i)^2 + \text{std}(x_i)^2\) over all network parameters,
  we get
  \[
    \|\text{rms}(x_i)\|^2 \;=\; \|\text{mean}(x_i)\|^2 \;+\; \|\text{std}(x_i)\|^2
  \]
  \[
    \frac{\|\text{rms}(x_i)\|^2}{\|\text{mean}(x_i)\|^2} \;=\; 1 \;+\; \frac{\|\text{std}(x_i)\|^2}{\|\text{mean}(x_i)\|^2}
  \]
  
  \[
    \text{SNR}^2
    \;=\;
    \frac{\phi^2}{1 - \phi^2}
    \;\;\;\Longrightarrow\;\;\;
    \text{SNR}
    \;=\;
    \sqrt{\frac{\phi^2}{\,1 - \phi^2\,}}
    \;=\;
    \frac{\phi}{\sqrt{1 - \phi^2}}.
  \]
\end{proof}

\section{SNR details}

\begin{figure}[H]
 \centering
 \includegraphics[width=.9\textwidth]{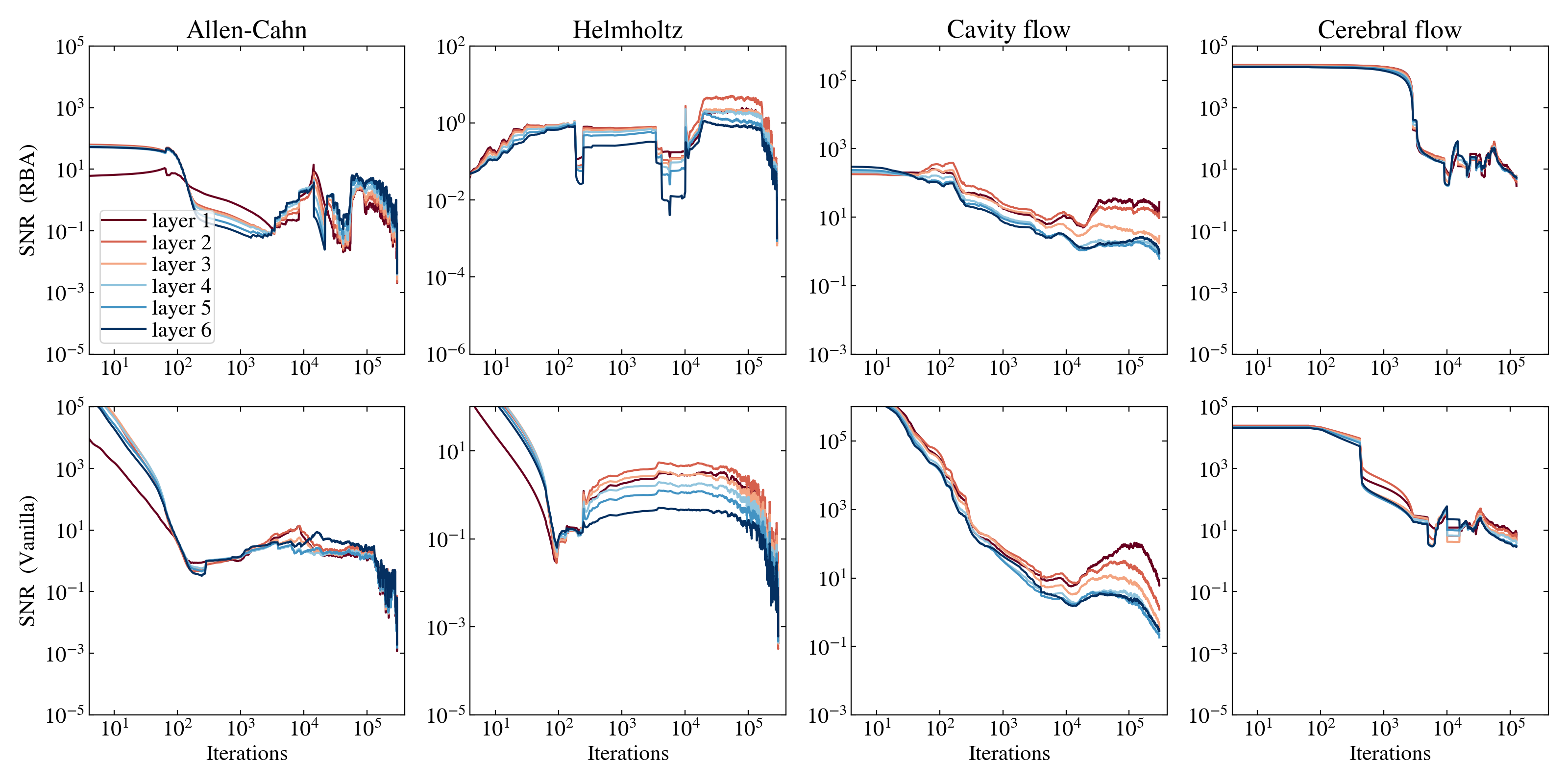}
 \caption[Layer-wise SNR]{ \textbf{Layer-wise SNR:} The SNR progression for each layer of the DNN follows a similar phase transition into DE (dashed lines). Moreover, Helmholtz and cavity have a decreasing SNR hierarchy from shallow to deep layers, which aligns with the IB theory. For Allen-Cahn and cerebral flow, the layers largely coincide, which could be attributed to minimal information-theoretic loss.}
 \label{p3:snr_layers}
\end{figure}

\begin{figure}[H]
 \centering
 \includegraphics[width=0.8\textwidth]{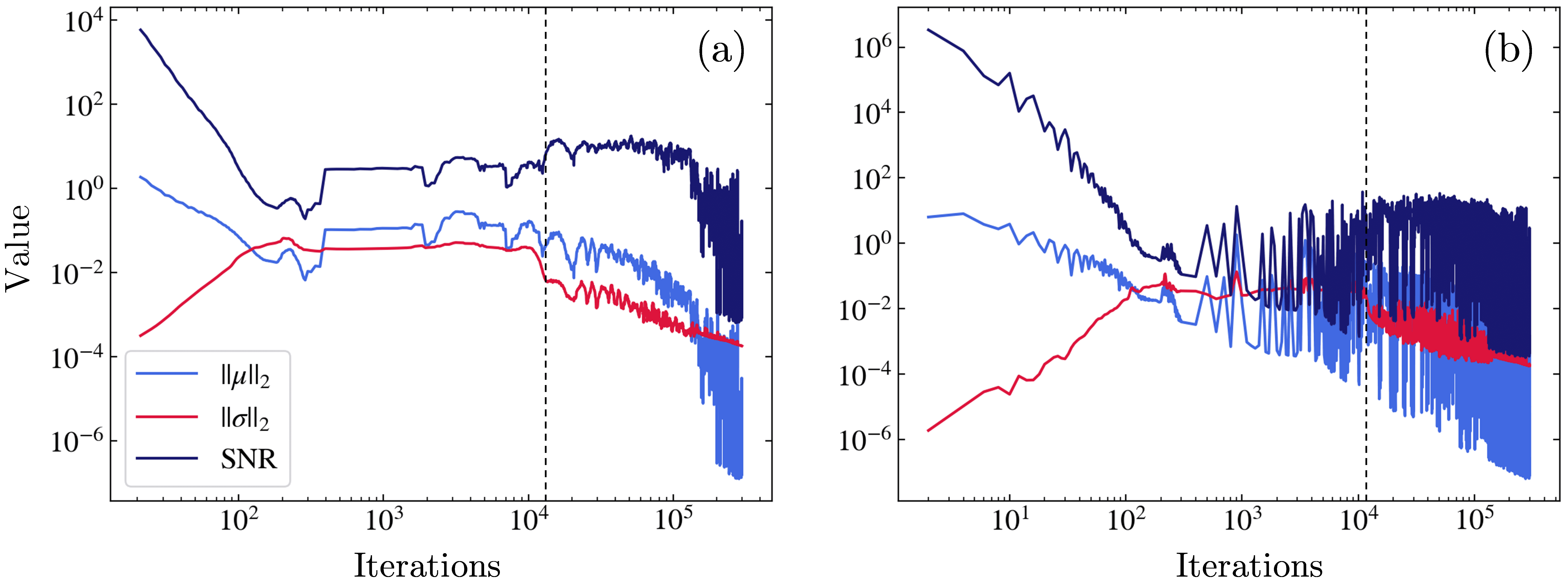}
 \caption[Signal and noise trajectories in SNR]{ \textbf{Signal and noise trajectories in SNR:} Both fitting and equilibrium (dashed lines) phases are characterized by an increased SNR. However, during fitting, the SNR starts high due to the low std of the batches since, at the beginning of training, the direction distinguishing each batch has not been determined (a). The increase of the SNR during DE is also accompanied by stability, as both the mean and std decrease at a similar rate. In the late training stages, the mean tends to zero (as the optimizer converges) while some gradient noise persists (a). Notably, after the phase transition, large oscillations are present in the non-smoothed curves (b). This implies a certain self-stabilization regime that can overcome local minima while maintaining a stable SNR, which may be attributed to the edge of stability, characterized by a highly non-convex but stable equilibrium.}
 \label{p3:noise}
\end{figure}

In this section, we present some additional results regarding the SNR curves. In Figure \ref{p3:snr_layers}, we show the layer-wise SNR where each layer seems to follow a similar pattern with all the phase transitions occurring at the same iteration, while decreasing SNR hierarchies are forming (shallow to deep layers), as observed in IB theory. The layers almost coincide for Allen-Cahn, since the information carried is almost identical. In Figure \ref{p3:noise}, we plot the mean and std curves for Allen-Cahn, which provides a clearer picture on why the SNR fraction follows this specific trend. During fitting, the gradient signal $\mu$ decreases from a high value, and the noise $\sigma$ increases (at a slower rate). In contrast, during the DE $\sigma$ suddenly drops (due to batch gradients becoming uniform), with both $\mu$ and $\sigma$ decreasing with a similar rate until the model converges.

\begin{figure}[H]
 \centering
 \includegraphics[width=.8\textwidth]{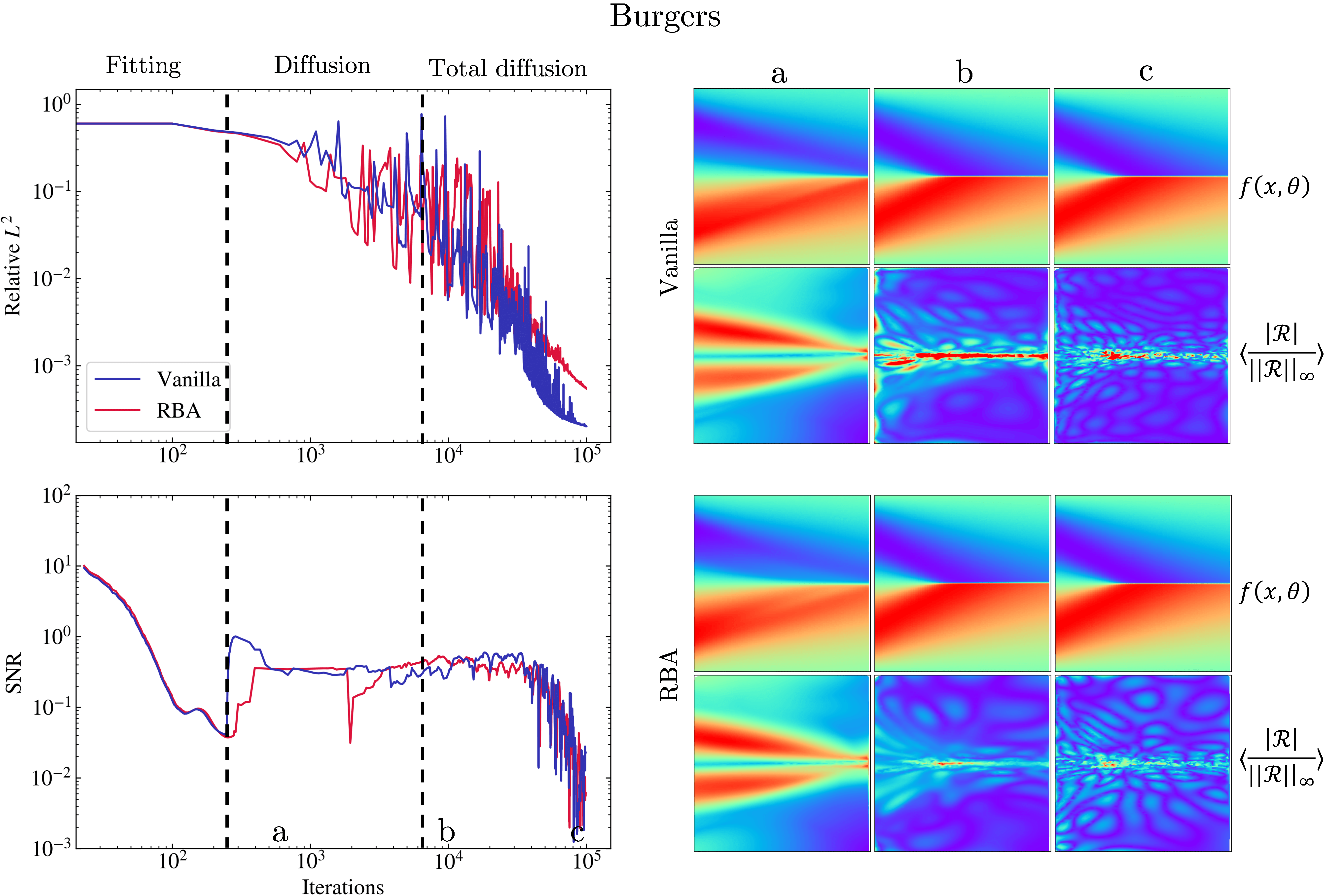}
 \caption[Burgers training dynamics]{ \textbf{Burgers training dynamics:} Both models follow similar $\text{SNR}$ paths, which undergo a high-to-low transition. Here, the $\text{SNR}$ remains below one during DE, probably due to the persistent steep gradient that causes a discontinuity on the centerline, hence increased stochasticity in the gradients. To make the residual homogeneity more visible for the vanilla case, we have limited the maximum range of their values (only for Burgers) since the high residuals at the discontinuity dominated the rest of the domain. Note that the Burgers equation is not considered a stiff problem, so the training quickly enters the equilibrium phase after about 500 iterations, unlike the other cases. Here, the RBA model performs a bit worse than vanilla since the weighting multipliers focus on a discontinuous region, which cannot be resolved. Furthermore, the tesst error convergence is characterized by large fluctuations due to the discontinuity, which is also reflected in the \text{SNR} order of magnitude.}
 \label{p3:res_BS}
\end{figure}

In the Burgers case of Fig. \ref{p3:res_BS}, the discontinuity limits the gradient homogeneity due to the steep gradient, which is always present, so although both the SNR and the residuals undergo a phase transition, the SNR does not become larger than one. The convergence here is non-monotonic, which again could be related to the “edge of stability” phase \cite{cohen2020gradient,damian2022self}. As expected, the more pronounced fluctuations are present due to the discontinuity. The vanilla PINN performs best here since the weighting multipliers of RBA focus on a discontinuous region, which cannot be resolved.

\section{Implementation Details}
{\bf Allen-Cahn Equation}

The general form of the 1D Allen-Cahn PDE is given by:
\begin{equation}
\frac{\partial u}{\partial t} = \epsilon\frac{\partial^2 u}{\partial x^2} + f(u),
\label{p3:AC}
\end{equation}
\noindent where $\epsilon > 0$ represents the diffusion coefficient, and $f(u)$ denotes a non-linear function of $u$. For this study, we have set $\epsilon = 10^{-4}$ and defined $f(u) = -5u^3 + 5u$. The initial condition is given by:
\begin{equation}
u(0, x) = x^2 \cos(\pi x), \quad \forall x \in [-1, 1],
\label{p3:IC_AC}
\end{equation}

To solve the Allen-Cahn PDE, we trained a physics-informed neural network (PINN) for $3\cdot10^5$ iterations using the standard Adam optimizer \cite{kingma2014adam} and an exponential learning rate scheduler, evaluated on 25600 collocation points. In general, the loss function $\mathcal{L}(\bm{x},\theta)$ is comprised of weighted terms to ensure adherence to the initial condition, boundary condition, and the PDE residual:
\begin{align*}
  \mathcal{L}(\bm{x},\theta) = \mathcal{L}_{ic} + \mathcal{L}_{bc} + \mathcal{L}_{\mathcal{R}},
\end{align*}
\noindent where $\mathcal{L}_{ic}=\langle (f(x_{ic}, \theta)-y_{ic})^2\rangle$ is the loss term for the initial condition, $\mathcal{L}_{bc}=\langle (f(x_{bc}, \theta)-y_{bc})^2\rangle $ is the boundary condition and $\mathcal{L}_{\mathcal{R}}=\langle \mathcal{R}(f(x, \theta))^2\rangle $ is the term enforcing the PDE residuals from Equation~\ref{p3:AC}. For Allen-Cahn, we only apply the $ic$ and residual terms. The RBA weights are applied solely for the PDE residual term $\mathcal{L}_{\mathcal{R}}$ \cite{anagnostopoulos2024residual}, as discussed in the main text.

{\bf Helmholtz Equation}

The 2D Helmholtz PDE is expressed as follows:
\begin{equation}
\frac{\partial^2 u}{\partial x^2} + \frac{\partial^2 u}{\partial y^2} + k^2u - q(x,y) = 0,
\label{p3:HM_eqn}
\end{equation}

\noindent where $q(x,y)$ is the forcing term defined as:
\begin{equation}
\begin{split}
q(x,y) = &- (a_1\pi)^2 \sin(a_1\pi x) \sin(a_2\pi y) \\
&- (a_2\pi)^2 \sin(a_1\pi x) \sin(a_2\pi y) \\
&+ k \sin(a_1\pi x) \sin(a_2\pi y),
\end{split}
\label{p3:HM:FT}
\end{equation}

\noindent leading to the analytical solution $u(x,y) = \sin(a_1\pi x) \sin(a_2\pi y)$ \cite{mcclenny2020self}. The boundary conditions are defined as:
\begin{equation}
u(-1,y) = u(1,y) = u(x,-1) = u(x,1) = 0,
\label{p3:HM:BC}
\end{equation}

For this study, we set $a_1 = 6$, $a_2 = 6$, and $k = 1$. The loss function for this problem is given by:
\begin{align*}
\mathcal{L}(\bm{x},\theta) = \mathcal{L}_{bc}(\bm{x},\theta) + \mathcal{L}_{\mathcal{R}}(\bm{x},\theta)
\end{align*}

\noindent Similar to the previous case, RBA is only applied for the PDE residual term $\mathcal{L}_{\mathcal{R}}$. The PINN is trained for $3 \cdot 10^5$ iterations with the Adam optimizer and an exponential learning rate scheduler on a sample space of 25,600 collocation points. We note that to reduce noise for the RBA case, for each multiplier $\lambda_i$ the moving average $\frac{|\mathcal{R}_i|}{\|\mathcal{R}_i\|_{\infty}}$ is modified as $\frac{|\mathcal{R}_i|-\|\mathcal{R}_i\|_{-\infty}}{\|\mathcal{R}_i\|_{\infty}-\|\mathcal{R}_i\|_{-\infty}}$, so that the lower bound is always 0. This did not affect the performance compared to the original definition of RBA and was only done for better visualization.

{\bf Lid-driven cavity flow}

The governing physical laws for the lid-driven cavity problem are encapsulated by the 2D steady Navier-Stokes (NS) equations, represented by the momentum equations:
\begin{align}
(\mathbf{v}\cdot\mathbf{\nabla})\mathbf{v} = -\mathbf{\nabla} p + \frac{1}{Re} \nabla^2\mathbf{v},
\label{p3:Mom}
\end{align} 
and the continuity equation:
\begin{align}
\mathbf{\nabla}\cdot\mathbf{v} = 0,
\label{p3:Mass}
\end{align} 
where $\mathbf{v}(\bm{x}) = (u(\bm{x}),v(\bm{x}))$ represents the velocity field within the cavity, $p$ denotes the pressure, $Re = 1000$ is the Reynolds number, and $\bm{x} = (x,y) \in \Omega = (0,1) \times (0,1)$, with $\Omega$ denoting the two-dimensional cavity domain \cite{wang2021understanding}. The boundary conditions are specified as:
\begin{equation}
\textbf{v}(\bm{x}) = \begin{cases}
(1,0) & \text{if } \bm{x} \in \Gamma_1,\\
(0,0) & \text{if } \bm{x} \in \Gamma_2,
\end{cases}
\end{equation}
where $\Gamma_1$ is the top boundary and $\Gamma_2$ encompasses the sides and bottom of the cavity. For the top boundary condition, the edges were smoothed to avoid singularities forming at the points where $\Gamma_1$ coincides with $\Gamma_2$, hence creating a well-posed problem. We further simplify the problem by utilizing the stream function $\psi$, where $\mathbf{v} = \nabla \times \psi$, and approximate $u, v,$ and $p$ as follows:
\begin{align}
u(x,y) &= \frac{\partial \psi_{NN}}{\partial y}(x,y),\\
v(x,y) &= -\frac{\partial \psi_{NN}}{\partial x}(x,y),\\
p(x,y) &= p_{NN}(x,y),
\end{align}
\noindent where $\psi_{NN}$ and $p_{NN}$ are neural network outputs for the stream function and pressure, respectively. This approximation inherently satisfies the continuity equation, $\nabla \cdot \mathbf{v} = \nabla \cdot (\nabla \times \psi) = 0$, allowing us to focus on the momentum equations (\ref{p3:Mom}). The total loss function is defined as:
\begin{equation}
\mathcal{L} = \mathcal{L}_{bc} + \mathcal{L}_{ns},
\end{equation}
where the combined loss terms for boundary conditions ($\mathcal{L}_{bc}$) and Navier-Stokes equations ($\mathcal{L}_{ns}$) are given by:
\begin{align}
\mathcal{L}_{bc} &= \sum_{b=1}^{2}\langle (\mathcal{R}_{h,b})^2\rangle_h,\\
\mathcal{L}_{ns} &= \langle (\lambda_{j} \cdot \sum_{n=1}^{2} |\mathcal{R}_{j,n}|)^2 \rangle_j,
\end{align}
\noindent here, $\mathcal{R}_{h,b}$ and $\mathcal{R}_{j,n}$ denote the residuals for boundary conditions and Navier-Stokes equations at collocation points $h$ and $j$, respectively. Finally, $\lambda_{j}$ represents the RBAs applied to PDE residual points.

To address this system of PDEs, a PINN was trained for $3 \times 10^5$ with Adam, coupled with an exponential learning rate scheduler. The training process was again applied on a set of 25,600 collocation points.

{\bf 3D Cerebrospinal fluid flow}

\begin{figure}[H]
 \centering
 \includegraphics[width=1.0\textwidth]{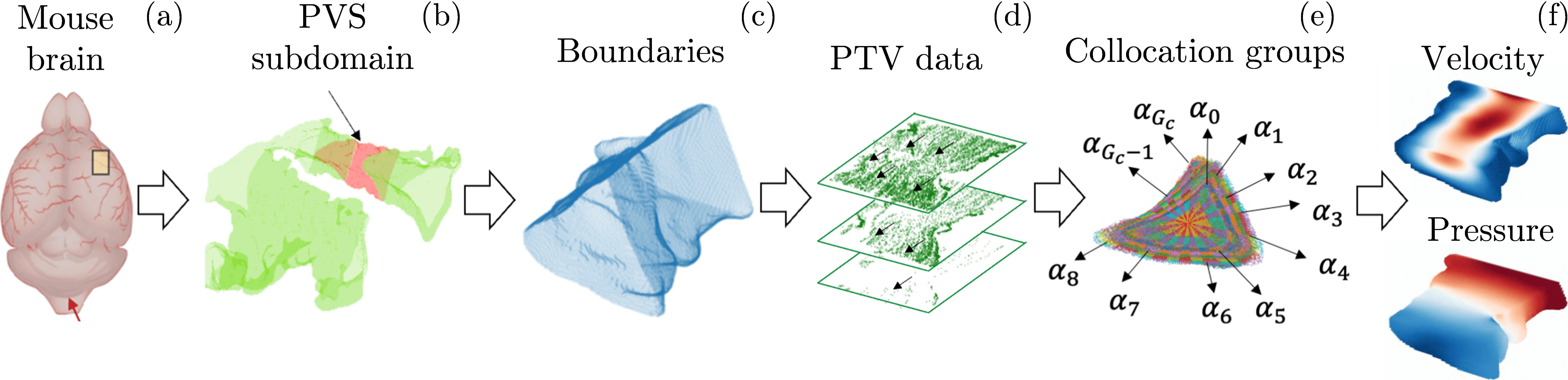}
 \caption[CSF flow reconstruction]{\textbf{CSF flow reconstruction:} (a) Tracer particles of one-micron size are introduced into the brain of a murine to measure the CSF flow within the perivascular spaces, while the velocities are measured via particle tracking velocimetry (PTV). (b) The perivascular space geometry is then derived from a volume scan and (c) the 3D boundaries for a subdomain are defined. (d) The velocity data is extracted from particle trajectories on three planes. (e) The collocation points are then created based to their spatial and temporal coordinates, organized into $G_{c}$ training groups ($\alpha_i$). (f) Finally, the 3D high-resolution velocity and pressure fields are inferred by a PINN.}
 \label{p3:Brain_sum}
\end{figure}

In this real-world case, we reconstruct the 3D high-resolution CSF transient velocity and pressure fields from experimental data collected within the perivascular spaces (PVS) of a murine's brain \cite{boster2023artificial}. The internal flow velocity measurements for a subdomain were obtained using particle tracking velocimetry (PTV), while non-slip boundary conditions were applied on the boundaries of the derived geometry which was obtained via a volume scan (Fig. \ref{p3:Brain_sum}(a-c)). Subsequently, 2D velocity data points from three different planes were collected, as illustrated in Figure ~\ref{p3:Brain_sum}(d). Furthermore, to enhance the temporal resolution of the PTV dataset, \cite{boster2023artificial} grouped the PTV data from multiple cardiac cycles into a single (phase-merged) cardiac cycle of $T=0.303 s$. As described in \cite{anagnostopoulos2024residual}, we partition our $5\times10^6$ collocation points and $5\times10^5$ boundary points (BCs) into \(5\times10^4\) PDE and \(5\times10^3\) boundary training groups, respectively, based on their spatial and temporal coordinates (see Figure~\ref{p3:Brain_sum}(e)). Then, we obtain our mini-batches by selecting one point from each training group. We then train the model using ordered mini-batches for the BCs and the PDE system. However, the entire dataset is used for the PTV data in each iteration. Finally, we reconstruct the 3D velocity and pressure fields for a full cardiac cycle (Fig. \ref{p3:Brain_sum}(f)). The final 2D plots shown in the Results section are produced by taking the rms of the residuals for the planes of Figure \ref{p3:Brain_sum}(d), projected on the central plane and averaged over three equidistant time-steps within [0, 0.303] s.


The CSF flow is governed by the incompressible 3D transient NS equations.
\begin{align}
  \frac{\partial\mathbf{v}}{\partial t} + (\mathbf{v}\cdot\mathbf{\nabla})\mathbf{v} &=-\mathbf{\nabla} p+\frac{1}{Re} \nabla^2\mathbf{v} \\
  \mathbf{\nabla}\cdot\mathbf{v}&=0\text{.}
\end{align} 

Following \cite{boster2023artificial,anagnostopoulos2024residual}, we use a neural network to obtain high-resolution velocities and pressure fields (Fig.~\ref{p3:Brain_sum}(f)) that follow the flow defined by the  PTV data and the BCs. In particular, we approximate $u,v,w$ and $p$ as follows:

\begin{align}
    u&=u_su_{NN}(t,x,y,z)\\
    v&=v_sv_{NN}(t,x,y,z)\\
    w&=w_sw_{NN}(t,x,y,z)\\
    p&=p_sp_{NN}(t,x,y,z)\text{,}
\end{align}

\noindent where $t,x,y,z$ are the inputs (i.e., time and position), $u_{NN},v_{NN},w_{NN},p_{NN}$ are the neural network outputs and $(u_{s},v_{s},w_{s},p_{s})$  are scalars ensuring that the network outputs have the same order of magnitude. For this case, we set $v_s=1$ (i.e., axial velocity), $u_s=w_s=0.1$ and $p_s=100$. 

 We apply global weights ($w_{\beta=\{b,d,n\}}$) to balance the contribution of the multiple loss terms (NS, PTV data, and BCs). Additionally, we apply RBA weights to control the interplay inside each loss term. We define the total loss function as  $\mathcal{L}=\mathcal{L}_{bc}+\mathcal{L}_{d}+\mathcal{L}_{ns}$, where the loss terms for BCs ($\mathcal{L}_{bc}$), PTV data ($\mathcal{L}_{d}$) and NS ($\mathcal{L}_{ns}$) are defined as follows:
\begin{align}
  \mathcal{L}_{bc}& = \sum_{b=1}^{3}w_{b}\langle (\lambda_{h} \cdot \mathcal{R}_{h,b})^2\rangle_h\\
  \mathcal{L}_{d}& = \sum_{d=1}^{2}w_d\langle(\lambda_{i} \cdot \mathcal{R}_{i,d})^2\rangle_i\\
  \mathcal{L}_{ns}& = \sum_{n=1}^{4}w_{n}\langle(\lambda_{j} \cdot \mathcal{R}_{j,n})^2\rangle_j\text{,}
\end{align}

\noindent where $\mathcal{R}_{h,b}, \mathcal{R}_{i,d}, \mathcal{R}_{j,n}$ are the BC, data, and NS residuals for points $h,i$, and $j$, $w_{b}, w_d, w_{n}$ are global weights multiplying each loss term, $\lambda_{h}$, $\lambda_{j}, \lambda_{i}$ are the RBA assigned for the BCs, NS and PTV collocation groups. The BCs and NS are averaged over their corresponding batch sizes, while the data terms are averaged over the PTV points.

To stabilize the training process and speed up convergence, we reparameterize our network using weight normalization (WN)\cite{salimans2016weight} and 
train the model for  $1.25\cdot10^5$ iterations using Adam optimizer \cite{kingma2014adam} with decaying learning rate.

{\bf Burgers' Equation}

The Burgers' equation is defined as:
\begin{equation}
  u_t + uu_x = \nu u_{xx},
  \label{p3:eq:Burgers}
\end{equation}
\noindent where $u$ represents the velocity field, subject to the dynamic viscosity $\nu=1/(100\pi)$. The initial condition and boundary conditions are described as follows:
\begin{align}
  u(0, x) = -\sin(\pi x), \quad \forall x \in \Omega,\\
  u(t, -1) = u(t, 1) = 0, \quad \forall t \geq 0,
  \label{p3:Burgers_BC}  
\end{align}

\noindent defined over the domain $\Omega = (-1,1) \times (0,1)$, where $\bm{x} = (x, y)$ signifies the spatial coordinates. The loss function is:

\begin{equation*}
  \mathcal{L} = \mathcal{L}_{bc}+ \mathcal{L}_{\mathcal{R}}
\end{equation*}
\noindent where $\mathcal{L}_{bc}$ and $\mathcal{L}_{\mathcal{R}}$ denotes the loss functions from the boundary conditions and residual points to satisfy equations~\ref{p3:Burgers_BC} and~\ref{p3:eq:Burgers} respectively. We minimize the total loss using Adam optimizer with exponential learning rate scheduler for $10^5$ iterations. For this problem, we use 10000 collocation points.


\end{document}